\documentclass[10pt,letterpaper]{article}
\usepackage{times}

\usepackage[T1]{fontenc}
\usepackage[authoryear,round]{natbib}

\usepackage{amsmath,amssymb,amsthm}

\usepackage{amsmath,amsfonts,bm}

\def\eqref#1{equation~\ref{#1}}

\def\1{\bm{1}}

\def\vzero{{\bm{0}}}

\def\vtheta{{\bm{\theta}}}
\def\va{{\bm{a}}}
\def\vb{{\bm{b}}}

\def\ve{{\bm{e}}}

\def\vh{{\bm{h}}}

\def\vr{{\bm{r}}}
\def\vs{{\bm{s}}}

\def\vu{{\bm{u}}}

\def\vw{{\bm{w}}}
\def\vx{{\bm{x}}}
\def\vy{{\bm{y}}}
\def\vz{{\bm{z}}}

\def\mI{{\bm{I}}}

\def\mP{{\bm{P}}}

\def\mW{{\bm{W}}}

\DeclareMathAlphabet{\mathsfit}{\encodingdefault}{\sfdefault}{m}{sl}
\SetMathAlphabet{\mathsfit}{bold}{\encodingdefault}{\sfdefault}{bx}{n}

\def\gG{{\mathcal{G}}}

\newcommand{\E}{\mathbb{E}}

\newcommand{\R}{\mathbb{R}}

\usepackage{graphicx,wrapfig}
\usepackage{longtable,array,booktabs}
\usepackage{caption,capt-of}
\usepackage{xcolor,colortbl}

\usepackage{titletoc}
\usepackage{url}
\usepackage[
    colorlinks=true,
    linkcolor=red,
    citecolor=blue,
    urlcolor=blue
]{hyperref}

\definecolor{customBlue}{RGB}{135,206,250}

\date{}

\theoremstyle{plain} 

\theoremstyle{remark}
\newtheorem{remark}{Remark}

\theoremstyle{definition} 

\newcommand{\mstd}[2]{#1_{\pm #2}}
\title{Permutation-Equivariant Flow Matching for Alignment-Free Neural Weight Generation}

\author{
\begin{tabular}{@{}l@{}}
\textbf{Arkadi Piven$^{1}$\thanks{Corresponding author:
\texttt{arkadi.piven@campus.technion.ac.il}.} \quad
Yam Eitan$^{1}$ \quad
Guy Bar-Shalom$^{1}$ \quad
Fabrizio Frasca$^{1}$} \\
\textbf{Daniel Cremers$^{2,3}$ \quad
Thomas Dag\`es$^{2,3}$ \quad
Ron Kimmel$^{1}$ \quad
Haggai Maron$^{1,4}$} \\[2pt]
{\small $^{1}$Technion -- Israel Institute of Technology
\quad $^{2}$Technical University of Munich} \\
{\small $^{3}$Munich Center for Machine Learning
\quad $^{4}$NVIDIA}
\end{tabular}
}

\RequirePackage{geometry}
\makeatletter
\renewcommand{\normalsize}{%
  \@setfontsize\normalsize{10pt}{11pt}%
  \abovedisplayskip 7pt plus 2pt minus 5pt
  \belowdisplayskip \abovedisplayskip
  \abovedisplayshortskip 0pt plus 3pt
  \belowdisplayshortskip 4pt plus 3pt minus 3pt
}
\renewcommand{\small}{\@setfontsize\small{9pt}{10pt}}
\renewcommand{\footnotesize}{\@setfontsize\footnotesize{9pt}{10pt}}
\renewcommand{\large}{\@setfontsize\large{12pt}{14pt}}
\renewcommand{\Large}{\@setfontsize\Large{14.4pt}{16pt}}
\renewcommand{\LARGE}{\@setfontsize\LARGE{17.28pt}{20pt}}

\renewcommand{\section}{\@startsection{section}{1}{\z@}%
  {-2ex plus -.5ex minus -.2ex}{1.5ex plus .3ex minus .2ex}%
  {\large\bfseries\raggedright}}
\renewcommand{\subsection}{\@startsection{subsection}{2}{\z@}%
  {-1.8ex plus -.5ex minus -.2ex}{.8ex plus .2ex}%
  {\normalsize\bfseries\raggedright}}
\renewcommand{\subsubsection}{\@startsection{subsubsection}{3}{\z@}%
  {-1.5ex plus -.5ex minus -.2ex}{.5ex plus .2ex}%
  {\normalsize\bfseries\raggedright}}
\renewcommand{\paragraph}{\@startsection{paragraph}{4}{\z@}%
  {1.5ex plus .5ex minus .2ex}{-1em}{\normalsize\bfseries}}

\renewcommand{\@maketitle}{%
  \vbox{\hsize\textwidth
    \vskip -1.5em
    \centering
    {\LARGE\normalfont\@title\par}%
    \vskip 1em
    {\normalsize\begin{tabular}[t]{c}\@author\end{tabular}\par}%
    \vskip .3in minus .1in
  }%
}
\makeatother

\normalsize
\renewenvironment{abstract}{%
  \par\vskip .075in
  \centerline{\large\bfseries Abstract}%
  \vspace{.5ex}%
  \normalsize
  \begin{list}{}{%
    \setlength{\leftmargin}{36pt}%
    \setlength{\rightmargin}{36pt}%
    \setlength{\listparindent}{0pt}%
    \setlength{\itemindent}{0pt}%
    \setlength{\topsep}{4pt plus 1pt minus 2pt}%
    \setlength{\partopsep}{1pt plus .5pt minus .5pt}%
    \setlength{\parsep}{2pt plus 1pt minus .5pt}%
  }%
  \item\relax
}{%
  \end{list}\par\vskip 1ex
}

\date{}
\begin{document}

\maketitle

\vspace{-1.5em}
\begin{abstract}
\vspace{-1.0em}
A 
trained
neural network can be represented by a parameter vector in 
high dimensions.
Learning distributions over these vectors enables the
generation of new models across various tasks and architectures. A central challenge is permutation symmetry: permuting hidden neurons
can produce distant parameter vectors representing the same function.
This introduces variations that a generative model must
account for when learning from trained networks. Existing methods typically address this 
using networks derived from a common base model or 
costly approximate neuron alignment.
We instead parameterize a flow-matching velocity field with a permutation-equivariant Graph Meta Network, enabling direct learning from independently trained networks without alignment. Extensive experiments show that our method closely reproduces the joint statistics of accuracy, functional similarity, and weight similarity of independently trained collections, providing evidence of generation beyond checkpoint memorization. A single conditional model also generates task-specific networks 
on
heterogeneous architectures and generalizes to unseen hidden-width configurations.
On a tabular domain-shift task, intermediate conditioning produces individual networks with performance comparable to logit ensembles across both domains. Taken together, our results show how permutation equivariance enables 
learning
from diverse collections of independently trained networks without permutation alignment.

\end{abstract}

\section{Introduction}

Trained network weights have emerged as a new data modality~\citep{unterthiner2020predicting,eilertsen2020classifying,schurholt2021self,navon2023equivariant, navon2024equivariant, lim2024graph, putterman2024learning, gelberg2026gradmetanet, dayan2026expressive, knyazev2025accelerating, tran2025equivariant,kofinas2024graph,kalogeropoulos2024scale}. 
Weight-space generative modeling
promises a paradigm shift: 
rather than optimize neural networks,
we could sample 
them
directly from a learned distribution.
Weight generation, rather than optimization, 
may unlock powerful capabilities, most notably conditional generation, allowing us to synthesize networks tailored to specific datasets, architectures, or domains~\citep{wang2026position,han2026survey}. 
Despite this potential, learning the distribution of independently trained neural networks remains a challenge.

\begin{wrapfigure}{r}{0.42\textwidth}
    \centering
    \vspace{-1.7em}
    \includegraphics[width=\linewidth]{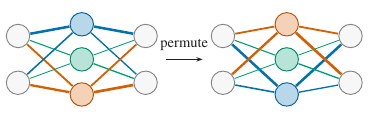}
    \vspace{-1.5em}
    \caption{Hidden-unit permutations preserve the network function.
    Colors track each hidden unit and its incident weights.}
    \vspace{-1em}
    \label{fig:permutation}
\end{wrapfigure}

Ideally, we wish to train a generative model on large-scale 
network collections,
spanning diverse tasks and architectures, to capture the distribution of diverse solutions. 
However, two functionally identical networks can 
have
different weight matrices due to 
arbitrary 
hidden-neuron ordering
(Figure~\ref{fig:permutation}). 
This is a practical concern as repositories like Hugging Face~\citep{wolf2020transformers} host 
diverse
trained networks without a common ordering or shared initialization. 
As such, weight-space symmetries, transformations that change 
weights 
yet
preserve
the function,
especially permutation symmetries, 
are ubiquitous
and thus
form a fundamental obstacle in weight-space generation. 

To 
avoid
this issue, prior 
works
alter
the 
network collections used for training.
Post-hoc approaches approximately align compatible 
training
networks to a common reference~\citep{ainsworth2022git}. 
A priori methods instead 
create artificially
the training collection 
so that alignment is 
not
needed, 
typically
by sampling checkpoints along the optimization trajectory of a single network~\citep{wang2024neural, wang2026scaling, erkocc2023hyperdiffusion}. 
%
These strategies require either compatible architectures for
alignment or control over how the training collection is
constructed, restricting their applicability to existing
repositories of independently trained networks.
%
%
%
Many generators impose 
further 
limitations
through fixed-dimensional parameterizations, 
restricting
their direct reuse across architectures~\citep{wang2024neural, wang2026scaling,erkocc2023hyperdiffusion,schurholt2022hyper, schurholt2024sane, gupta2026deepweightflow}.
Moreover, sample efficiency and memorization remain challenges even within these restricted settings~\citep{zeng2026generative,navon2023equivariant, navon2024equivariant, shamsian2024improved}.

\begin{figure}[t]
    \centering
    \includegraphics[width=\textwidth]{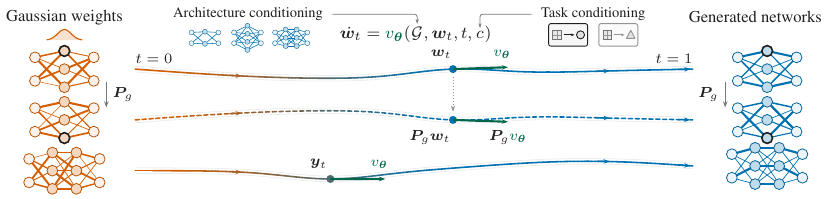}
    \caption{\textbf{Method overview.}
    A shared velocity field $v_{\vtheta}(\gG,\vw_t,t,c)$,
    trained on unaligned networks, generates weights conditioned
    on architecture $\gG$ and task $c$.
    States $\vw_t$ and $\vy_t$ belong to different architectures.
    A hidden-unit permutation $\mP_g$ permutes the trajectory
    and its velocities, producing functionally equivalent
    networks (upper pair).}
    \label{fig:method_overview}
\end{figure}

We advocate a switch to equivariant tools, which are natural for 
symmetries.
Following recent advances in graph metanetworks
~\citep{lim2024graph, knyazev2025accelerating, kofinas2024graph, kalogeropoulos2024scale}, we represent neural networks as graphs, adopt Flow Matching (FM)~\citep{lipman2023flow}, and parameterize the velocity field with a Graph Meta Network (GMN) designed to be equivariant with weight-space permutations (Figure~\ref{fig:method_overview}). We motivate this inductive bias by showing that permutation-invariant initialization and equivariant training yield a permutation-equivariant optimal FM velocity field.
On 
up to $100\mathrm{K}$ independently trained networks, our model generates networks with joint accuracy and similarity statistics 
much
closer to the training 
network distribution
than 
existing works.
We further show that our method can interpolate across domain shifts, process heterogeneous architectures, and generate weights at unseen hidden widths.

Our main contributions can be summarized as follows:
\begin{enumerate}
    \item \textbf{Alignment-free generation beyond memorization.} We parameterize a flow-matching velocity field with a permutation-equivariant GMN to learn directly from independently trained networks without alignment. Our proposed joint quality measure shows that generated networks closely match independently trained networks in performance and similarity, providing evidence against checkpoint memorization.
    \item \textbf{Heterogeneous architecture generation.}
        A single flow generates weights across heterogeneous
        architectures and generalizes to unseen widths, without
        retraining or architecture-specific changes to the generator.
     \item \textbf{Conditional weight-space traversal.}
        We demonstrate conditional generation across diverse tasks
        and domain shifts, showing that our flow generalizes to
        unseen conditions and synthesizes single unified models
        on par with logit ensembling.

\end{enumerate}

\section{Preliminaries}
\label{preliminaries}
\noindent\textbf{Weight space permutation symmetries. }
\label{subsec:symmetries}
Consider a multilayer perceptron (MLP) $F(\vx;\vw)$ with $L$ hidden layers with widths $d_0,\ldots,d_{L+1}$, where $\vx\in\R^{d_0}$
is the input and $\vw$ is the vectorized network's weights and
biases. 
The weight and bias of layer $\ell$ are given by $\mW_\ell\in\R^{d_\ell\times d_{\ell-1}}$ and 
$\vb_\ell\in\R^{d_\ell}$.
%
For each hidden
layer, let $\mP_\ell$ be a permutation matrix, and fix
$\mP_0=\mI_{d_0}$ and $\mP_{L+1}=\mI_{d_{L+1}}$ to identity matrices to preserve
the input and output coordinates. 
Changing parameters to
\begin{equation}        
    \mW_\ell'=\mP_\ell\mW_\ell\mP_{\ell-1}^{\top}
    \quad\text{and}\quad
    \vb_\ell'=\mP_\ell\vb_\ell,
    \qquad 
    \text{for }\; \ell=1,\ldots,L+1,
\end{equation}
reorders 
hidden neurons 
and
their
incoming and
outgoing weights. 
As
elementwise activations commute
with permutations,
distinct parameters $\vw'$ can represent the same function~\citep{hecht1990algebraic, navon2023equivariant}
\begin{equation}
    F(\vx;\vw')=F(\vx;\vw),
    \qquad \forall \vx \in\mathbb{R}^{d_0}.
    \label{eq:permutation_invariance}
\end{equation}
The same goes for
convolutional 
networks 
and
Transformers~\citep{kofinas2024graph,lim2024graph}.

\noindent\textbf{Graph metanetworks.}
Graph metanetworks 
view
a neural network as a \emph{parameter graph} $\gG$: nodes correspond to neurons or channels, and edges carry the associated weights~\citep{lim2024graph, kofinas2024graph, kalogeropoulos2024scale, bar2026graph}. 
Added
positional encodings 
identify structural roles, 
e.g.,
layer membership. A Graph Neural Network (GNN) whose message-passing layers naturally respect weight-space symmetries 
processes
this graph and 
outputs either edge features (e.g., new weights) or a global output (e.g., the network's accuracy on a dataset). The resulting metanetwork is equivariant to weight-space permutations, and its shared graph operations allow one model to process heterogeneous architectures.

\noindent\textbf{Flow Matching.}
Flow Matching (FM)~\citep{lipman2023flow} learns a velocity
field $v_{\vtheta}$
via
a neural network with
parameters $\vtheta$, to transport
a source distribution $p_0$ to target $p_1$.
With independent
coupling and linear interpolation: drawing
$\vw_0\sim p_0$, $\vw_1\sim p_1$, and
$t\sim\mathcal{U}(0,1)$ independently, we define $\vw_t=(1-t)\vw_0+t\vw_1$. The network is trained by minimizing
the flow matching objective
\begin{equation}
\label{fm_objective}
    \mathcal{L}_{\mathrm{FM}}(\vtheta)
    =
    \E_{t,\vw_0,\vw_1}
    \left[
        \left\|
            v_{\vtheta}(\vw_t,t)-(\vw_1-\vw_0)
        \right\|_2^2
    \right].
\end{equation}
At inference, we draw $\vw_0\sim p_0$ and integrate
the following learned velocity field
from $t=0$ to $1$
\begin{equation}
    \dot{\vw}_t=\frac{d\vw_t}{dt}=v_{\vtheta}(\vw_t,t).
\end{equation}

\section{Related Efforts}
Early work showed that neural networks can learn to generate other neural networks by producing their weights~\citep{ha2016hypernetworks, bertinetto2016learning, jia2016dynamic}. Building on this, many efforts were made toward modeling the distributions of the weights themselves.

\noindent\textbf{Weight-Space Generative Models.}
Hyper-Representations and SANE use autoencoders to learn
latent representations for weight
generation~\citep{schurholt2022hyper,
schurholt2024sane}.
P-diff~\citep{wang2024neural} combines an autoencoder with
latent diffusion, a framework extended with dataset
conditioning in D2NWG~\citep{soro2025diffusion}.
DeepWeightFlow (DWF) operates directly in weight space,
with optional PCA compression~\citep{gupta2026deepweightflow}, while RPG scales
generation through parameter tokenization and recurrent
diffusion~\citep{wang2026scaling}.

\noindent\textbf{Post-Hoc Alignment.}
Post-hoc approaches align existing checkpoints before
training the generator.
SANE~\citep{schurholt2024sane} and
DWF~\citep{gupta2026deepweightflow} use canonicalization
methods such as Git Re-Basin~\citep{ainsworth2022git},
which approximately matches weights to a common reference.
\citet{erdogan2025geometric} explores permutation-equivariant
FM on MLP trajectory checkpoints, incorporating
Git Re-Basin into the training pipeline. For Git Re-Basin, matching assumes equivalent
architectures, limiting its direct applicability to
heterogeneous collections.

\noindent\textbf{A Priori Alignment.}
A priori methods 
artificially
reduce permutation variability
by
building
dependent networks.
P-diff~\citep{wang2024neural} collects checkpoints along
an optimization trajectory, a recipe also adopted in
D2NWG~\citep{soro2025diffusion}.
RPG~\citep{wang2026scaling} similarly collects fine-tuning
checkpoints from pretrained models.
NNiT~\citep{kim2026nnit} 
builds
its 
structurally aligned training collection
using
a Graph HyperNetwork with a convolutional decoder.
Such recipes assume control
on creating dependent networks,
limiting their 
applicability
to existing repositories of independently trained checkpoints.

\noindent\textbf{Weight-Space Memorization.}
\citet{zeng2026generative} found substantial memorization
in P-diff~\citep{wang2024neural},
Hyper-Representations~\citep{schurholt2022hyper},
HyperDiffusion~\citep{erkocc2023hyperdiffusion},
and G.pt~\citep{peebles2022learning}.
In the settings studied, they found that these methods largely reproduced
training checkpoints or their interpolations and did not
outperform simple weight perturbation or averaging
baselines in jointly achieving high performance and
novelty.

\section{Method}
\label{sec:main_method}

Our goal is to learn weight distributions from independently
trained networks without assuming a common hidden-unit
ordering.
We motivate our generator through the symmetry of these
distributions, then construct a shared equivariant velocity
field across architectures and tasks.

\noindent\textbf{Notation.}
We represent a network by its parameter graph
$\gG=(\mathcal{V},\mathcal{E})$, with 
nodes
$\mathcal{V}$,
edges
$\mathcal{E}$, and $D_{\gG}$ parameters.
Its vectorized weights and biases are 
$\vw\in\R^{D_{\gG}}$, while $c$ 
is
a task or domain
condition and $\vtheta$ 
is
the learned parameters of
the velocity network.
For each architecture, $G$ 
is
the group of
function-preserving hidden-unit permutations, acting on
$\vw$
through permutation matrices $\mP_g$.
A probability law $p$ is $G$-invariant if
$\mP_g\vw$ has the same law as $\vw\sim p$
for every $g\in G$. For each architecture $\gG$ and task or domain condition $c$,
our target is the conditional weight distribution
$p_1(\cdot\mid\gG,c)$.
We extend the FM velocity field introduced in
Section~\ref{preliminaries} to
$v_\vtheta(\gG,\vw_t,t,c)$, sharing its parameters across
architectures and tasks while keeping $\gG$ and $c$ fixed
along each trajectory.
We first analyze the flow for a fixed pair $(\gG,c)$,
suppressing this conditioning in the notation.

\subsection{Motivation}

We first ask what structure a weight-space generator should respect.
Two observations answer this: permutation symmetry is inherited by the
distribution of independently trained networks, and consequently by the
optimal flow-matching velocity field.

\noindent \textbf{Symmetry of the weight distribution.} \label{data_symmetric}
We first establish an important property of 
typical weight distributions. For a fixed architecture-task pair $(\gG,c)$, consider a
training procedure $r$ 
specifying
the dataset, loss,
optimizer, hyperparameters, and stopping rule. Let $T_r:\mathbb{R}^{D_{\gG}} \to \mathbb{R}^{D_{\gG}}$ map an initialization to its final checkpoint. An initialization distribution $p_{\mathrm{init}}$ then induces a weight distribution $p_1$ through $\vw_1=T_r(\vw_{\mathrm{init}})$, where $\vw_{\mathrm{init}}\sim p_{\mathrm{init}}$.
If
$p_{\mathrm{init}}$ is $G$-invariant and $T_r$ is equivariant
\begin{equation}
    T_r(\mP_g\vw_{\mathrm{init}})
    =\mP_g T_r(\vw_{\mathrm{init}}),
    \qquad \forall g\in G,
    \label{eq:commutation}
\end{equation}
then
$p_1$ is 
$G$-invariant: permuting 
trained
weights leaves the probability law unchanged.
Usual
layerwise initializations satisfy 
this
assumption. SGD and Adam respect permutation symmetries when per-step losses and stopping rules
are permutation-invariant. The 
result
also holds for mixtures of training recipes,
provided each 
induces a $G$-invariant distribution.
This invariance concerns the underlying weight distribution:
a finite training collection need not itself be invariant
(Appendix~\ref{appendix:proof_population}).

\noindent\textbf{Equivariance of the velocity field.}
Building on the equivariant-flow framework of
\citet{kohler2020equivariant}, we motivate our
parameterization through the symmetry of the optimal
flow-matching velocity field. The isotropic Gaussian prior
$p_0=\mathcal{N}(\vzero,\mI_{D_{\gG}})$ is
$G$-invariant, as is the target weight distribution
$p_1$ under the assumptions above. When both prior and target are $G$-invariant, and under independent draws $\vw_0\sim p_0, \vw_1 \sim p_1$ with a linear interpolation $\vw_t=(1-t)\vw_0+t\vw_1$, the marginal velocity field $u_t(\vw)=\E[\vw_1-\vw_0\mid\vw_t=\vw]$ that minimizes the FM objective~\ref{fm_objective}~\citep{lipman2023flow} is almost surely $G$-equivariant
(Appendix~\ref{appendix:proof_marginal}):
\begin{equation}
    u_t(\mP_g\vw)=\mP_g u_t(\vw),
    \qquad \forall g\in G,\quad \forall t\in[0,1].
    \label{eq:marginal_equivariance}
\end{equation}

Since under natural conditions the optimal velocity field is 
equivariant, restricting the
hypothesis class to equivariant fields is a 
justified
inductive bias
for
learning from finite training collections. Equivariant parameterization also 
nullifies
training-collection
alignment 
(see Section~\ref{sec:inert}).

\subsection{Equivariant Weight-Space Flow Matching}
\label{sec:method_gnn}
We now parameterize the shared conditional velocity field
$v_\vtheta(\gG,\vw_t,t,c)$ to satisfy the equivariance in
Equation~\ref{eq:marginal_equivariance} for each fixed
$(\gG,c)$.

\noindent\textbf{Graph representation.}
Following \citet{lim2024graph} and~\citet{kofinas2024graph}, we encode
the network parameters on a graph.
Shared encoders map node positional and layer information
into features $\vh_i^{(0)}$.
Edge features $\ve_{ij}^{(0)}$ combine encoded weights
with embeddings of the edge's positional information,
layer, and parameter type.
We also maintain a global feature $\vu^{(\ell)}$
for the graph at each layer $\ell$.
Positional encodings respect hidden-unit permutations:
interchangeable neurons within a layer share an encoding,
while fixed roles, such as input coordinates and output
classes, do not.

\begin{figure}[t]
    \vspace{-1em}
    \centering
    \includegraphics[width=\linewidth]{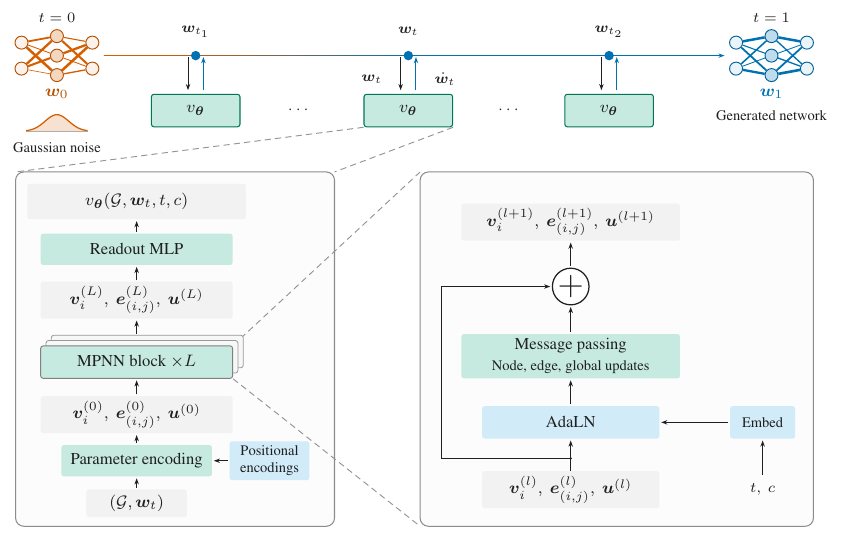}
    \caption{\textbf{Equivariant velocity network.}
    Top: an ODE solver transforms Gaussian weights into a
    functional network through repeated velocity evaluations.
    Bottom left: graph encoding, $L$ message-passing blocks,
    and parameter-velocity prediction.
    Bottom right: each block combines time- and task-conditioned
    AdaLN, message passing, and residual updates.}
    \label{fig:equivariant_flow_method}
\end{figure}

\noindent\textbf{Architecture.}
We parameterize $v_\vtheta(\gG,\vw_t,t,c)$ with a
message-passing GMN over the parameter graph $\gG$ (Figure~\ref{fig:equivariant_flow_method}).
The GMN uses $L$ message-passing blocks.
Let $\vh_i^{(\ell)}$, $\ve_{ij}^{(\ell)}$, and
$\vu^{(\ell)}$ denote the features of node $i$,
edge $(i,j)$, and the graph as a whole at layer $\ell$,
respectively.

Time $t$ and condition $c$ enter each block via
Adaptive Layer Normalization
(AdaLN)~\citep{peebles2023scalable} 
with
joint
embedding $\vz(t,c)$.
Bars, 
e.g., $\bar{\vh}$,
denote 
modulated features.
Let $\mathcal{N}(i)$ 
be
the neighbors of node $i$
and $\bigoplus$ a permutation-invariant 
aggregator.
For 
$0\le\ell\le L-1$,
the updates are:
\begin{align}
    \hspace{-0.6em}\left(
        \bar{\vh}_i^{(\ell)},
        \bar{\ve}_{ij}^{(\ell)},
        \bar{\vu}^{(\ell)}
    \right)
    &=
    \operatorname{AdaLN}_{\ell}\!\left(
        \left(
            \vh_i^{(\ell)},
            \ve_{ij}^{(\ell)},
            \vu^{(\ell)}
        \right),
        \vz(t,c)
    \right), \\
    \vh_i^{(\ell+1)}
    &=
    \vh_i^{(\ell)}
    +
    \operatorname{MLP}_{\mathrm{node}}^{(\ell)}
    \left(
        \bar{\vh}_i^{(\ell)},
        \bigoplus_{j\in\mathcal{N}(i)}
        \operatorname{MLP}_{\mathrm{msg}}^{(\ell)}
        \left(
            \bar{\vh}_i^{(\ell)},
            \bar{\vh}_j^{(\ell)},
            \bar{\ve}_{ij}^{(\ell)},
            \bar{\vu}^{(\ell)}
        \right),
        \bar{\vu}^{(\ell)}
    \right), 
    \\
    \ve_{ij}^{(\ell+1)}
    &=
    \ve_{ij}^{(\ell)}
    +
    \operatorname{MLP}_{\mathrm{edge}}^{(\ell)}
    \left(
        \bar{\vh}_i^{(\ell)},
        \bar{\vh}_j^{(\ell)},
        \bar{\ve}_{ij}^{(\ell)},
        \bar{\vu}^{(\ell)}
    \right), \\
    \vu^{(\ell+1)}
    &=
    \vu^{(\ell)}
    +
    \operatorname{MLP}_{\mathrm{global}}^{(\ell)}
    \left(
        \bigoplus_{i\in\mathcal{V}}\bar{\vh}_i^{(\ell)},
        \bigoplus_{(i,j)\in\mathcal{E}}\bar{\ve}_{ij}^{(\ell)},
        \bar{\vu}^{(\ell)}
    \right).
\end{align}

A shared readout MLP uses the final edge feature
$\ve_{ij}^{(L)}$ to predict the velocity of each weight.
The resulting field is equivariant to hidden-unit
permutations (Appendix~\ref{appendix:proof_parameterized}).

\noindent\textbf{Training.}
\label{sec:training_objective}
For each architecture-task pair $(\gG,c)$, 
$p_{\mathrm{data}}(\cdot\!\mid\!\gG,c)$ 
is
its empirical
weight distribution. 
Drawing 
independent 
$\vw_0\sim p_0^{(\gG)}
=\mathcal{N}(\vzero,\mI_{D_{\gG}})$,
$\vw_1\sim p_{\mathrm{data}}(\cdot\!\mid\!\gG,c)$,
and $t\sim\mathcal{U}(0,1)$, 
we
minimize
%
\begin{equation}
    \mathcal{L}(\vtheta)
    =
    \E_{(\gG,c),\,t,\,\vw_0,\,\vw_1}
    \left[
        \left\|
        v_\vtheta(\gG,\vw_t,t,c)
        -(\vw_1-\vw_0)
        \right\|_2^2
    \right],
    \label{eq:training_objective}
\end{equation}
with
$\vw_t=(1-t)\vw_0+t\vw_1$. 
The expectation
averages 
over
the training architecture-task pairs.
%
For sampling we use classifier-free guidance
(CFG)~\citep{ho2022classifier}, with
SDEdit-inspired refinement~\citep{meng2022sdedit} where specified,
alternating Gaussian corruption with flow integration.


\noindent\textbf{Alignment leaves the expected objective unchanged.}
\label{sec:inert}
For a permutation-equivariant velocity field, an invariant
prior, and independent coupling, applying a separate hidden-unit
permutation to each training network leaves the expected
flow-matching objective unchanged.
In particular, each target permutation can be absorbed into a corresponding
permutation of the noise, whose distribution is unchanged
(see proof in Appendix~\ref{appendix:proof_inert}).
Our model is therefore indifferent to how the training networks are
permuted: alignment preprocessing can neither help nor hurt it, so it
can be omitted rather than approximated at a cost
(Table~\ref{tab:alignment}).


\subsection{Evaluating Generation Beyond Memorization}
\label{sec:evaluation}

\begin{wrapfigure}{r}{0.42\textwidth}
    \centering
    \vspace{-2em}
    \includegraphics[width=\linewidth]{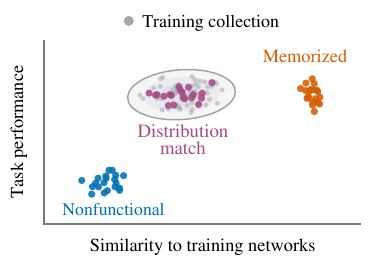}
    \vspace{-2em}
    \caption{
    \textbf{
    Generation joint evaluation
    (conceptual).}
    Memorized networks 
    perform well;
    nonfunctional networks 
    have low similarity.
    We seek agreement with
    training networks
    in both metrics.
    }
    \label{fig:evaluation_concept}
    \vspace{-2em}
\end{wrapfigure}

To assess generation beyond checkpoint
memorization~\citep{zeng2026generative}, we compare joint
task-performance, functional-similarity, and weight-similarity
statistics of generated and independently trained networks.
High performance can reflect copying, while low similarity
can reflect nonfunctional outputs
(Figure~\ref{fig:evaluation_concept}).
For a fixed task and architecture, 
$\mathcal{T}=\{\vw_i\}_{i=1}^{N}$ 
is
the training collection.

\noindent\textbf{Task performance and functional similarity.}
We use test accuracy, or positive-class ROC AUC,
as the task-performance score $S_{\mathrm{task}}(\vw)$,
and maximum error-IoU as functional similarity
$S_{\mathrm{func}}(\vw;\mathcal{T})$,
which
is 
a reference
measure of error-pattern
similarity in weight-space
generation~\citep{zeng2026generative,wang2024neural,
gupta2026deepweightflow}.
See Appendix~\ref{appendix:metrics} for a definition.

\noindent\textbf{Weight similarity.}
Functional similarity
may be fooled by small training weight perturbations.
We 
thus
also measure matched Weight Cosine Similarity
(WCS). Let $\mP_i$ be the hidden-unit permutation
found by aligning $\vw_i$ to $\vw$ using
Git Re-Basin (weight variant). We define
\begin{equation}
    \mathrm{WCS}(\vw;\mathcal{T})
    =
    \max_{\vw_i\in\mathcal{T}}
    \frac{\vw^\top\mP_i\vw_i}
         {\|\vw\|_2\|\mP_i\vw_i\|_2}.
\end{equation}

\noindent\textbf{Joint Wasserstein Similarity.}
Separate average 
scores mask
how performance and similarity vary together. 
Instead, we combine for each network the three scores:
task performance and functional 
and weight 
similarities.
We compare generated networks with those from the training collection, excluding self-comparisons when computing the latter's similarity scores.
We scale the three scores to balance their contributions,
applying the same factors from the fixed
training reference to generated and training networks.
Let $\{\vs_i\}_{i=1}^{M}$ be the resulting score vectors
of $M$ generated networks and
$\{\vr_j^{(b)}\}_{j=1}^{M}$ those of the $b$-th equally
sized 
sampled subset
of
the training 
reference.
To compare these sets of scores, we use the Wasserstein-2 distance,
which measures the cost to match their elements,
\begin{equation}
    W_2^{(b)}
    =
    \left[
        \min_{g\in S_M}
        \frac{1}{M}
        \sum_{i=1}^{M}
        \left\|
            \vs_i-\vr_{g(i)}^{(b)}
        \right\|_2^2
    \right]^{1/2},
    \label{eq:discrete_wasserstein}
\end{equation}
$S_M$ 
being
the permutation group on $M$ indices.
We introduce Joint Wasserstein Similarity (JWS) as
\begin{equation}
    \mathrm{JWS}
    =
    \max\left\{
        0,\,
        1-\frac{1}{B\sqrt{3}}
        \sum_{b=1}^{B}W_2^{(b)}
    \right\}.
    \label{eq:jws}
\end{equation}
Division by $B$ averages over reference samples,
while 
division
by $\sqrt{3}$ gives a root-mean-square
difference per coordinate.
Higher scores indicate closer agreement in the joint
performance and similarity statistics. 
This is our main quantitative generation score.
More details in Appendix~\ref{appendix:metrics}.

\section{Experiments}
\label{sec:experiments}
We evaluate unconditional and conditional generation on collections of up to
$100\mathrm{K}$ \emph{independently} trained image classifiers and tabular predictors.
Throughout, we compare against generators that rely on alignment or on
dependent training collections with emphasis on assessing memorization.

\subsection{Unconditional Generation}
\label{sec:unconditional_generation}

\noindent\textbf{Datasets.}
We use unaligned collections of $10\mathrm{K}$ independently
trained MNIST MLP classifiers with two hidden layers~\citep{lecun1998gradient}
and $50\mathrm{K}$ CIFAR-10 CNN classifiers with three convolutional
layers and a linear classifier~\citep{krizhevsky2009learning}.
All networks use distinct initialization seeds.

\noindent\textbf{Baselines.}
We 
evaluate
DWF~\citep{gupta2026deepweightflow},
P-diff~\citep{wang2024neural},
SANE~\citep{schurholt2024sane}, and MLP and
DiT~\citep{peebles2023scalable} velocity networks.
DWF is 
tested
with and without Git Re-Basin alignment.
SANE uses aligned weights, permutation augmentation, and KDE or Gaussian sampling.
Remaining baselines use unaligned weights.
Random weights and perturbed training checkpoints 
serve as
controls.
Experimental details and baseline sweeps are 
in
Appendix~\ref{appendix:unconditional_details}.










\begin{table}[t]
    \caption{Unconditional generation on MNIST and CIFAR-10.
    Accuracy (\%), Max IoU, and WCS report
    mean$_{\pm\mathrm{std}}$ across networks; JWS ($\uparrow$)
    averages over reference subsamples.
    Reference JWS $=1$ denotes an identity comparison.
    $^\dagger$ indicates aligned training weights. Perturbation reports the
    highest JWS over tested noise levels
    ($\sigma=0.25$ for MNIST; $0.10$ for CIFAR-10).
    GMN results include flow-guided refinement.
    Bold marks the highest JWS among evaluated methods.}
    \label{tab:unconditional}
    \centering
    \small
    \resizebox{\textwidth}{!}{%
    \setlength{\tabcolsep}{3.5pt}
    \begin{tabular}{@{}lcccccccc@{}}
        \toprule
        \textbf{Model}
        & \multicolumn{4}{c}{\textbf{MNIST}}
        & \multicolumn{4}{c}{\textbf{CIFAR-10}} \\
        \cmidrule(lr){2-5} \cmidrule(lr){6-9}
        & \textbf{Acc. (\%)} & \textbf{IoU}
        & \textbf{WCS} & \textbf{JWS} ($\uparrow$)
        & \textbf{Acc. (\%)} & \textbf{IoU}
        & \textbf{WCS} & \textbf{JWS} ($\uparrow$) \\
        \cmidrule(lr){2-4} \cmidrule(lr){5-5}
        \cmidrule(lr){6-8} \cmidrule(lr){9-9}
        Training ref.
        & $\mstd{95.9}{0.2}$ & $\mstd{0.559}{0.015}$
        & $\mstd{0.557}{0.010}$ & 1.00
        & $\mstd{79.6}{0.4}$ & $\mstd{0.609}{0.006}$
        & $\mstd{0.609}{0.013}$ & 1.00 \\
        \cmidrule(lr){2-4} \cmidrule(lr){5-5}
        \cmidrule(lr){6-8} \cmidrule(lr){9-9}
        Random init.
        & $\mstd{10.1}{2.4}$ & $\mstd{0.052}{0.002}$
        & $\mstd{0.138}{0.003}$ & 0.15
        & $\mstd{10.1}{1.7}$ & $\mstd{0.254}{0.012}$
        & $\mstd{0.255}{0.006}$ & 0.31 \\
        Perturbation
        & $\mstd{92.2}{2.2}$ & $\mstd{0.403}{0.073}$
        & $\mstd{0.971}{0.001}$ & 0.54
        & $\mstd{74.8}{1.7}$ & $\mstd{0.610}{0.047}$
        & $\mstd{0.995}{0.000}$ & 0.63 \\
        MLP
        & $\mstd{9.8}{3.7}$ & $\mstd{0.052}{0.002}$
        & $\mstd{0.355}{0.009}$ & 0.23
        & $\mstd{10.0}{1.6}$ & $\mstd{0.256}{0.012}$
        & $\mstd{0.146}{0.003}$ & 0.25 \\
        DiT
        & $\mstd{72.0}{6.5}$ & $\mstd{0.138}{0.032}$
        & $\mstd{0.555}{0.010}$ & 0.54
        & $\mstd{43.7}{5.5}$ & $\mstd{0.326}{0.025}$
        & $\mstd{0.553}{0.015}$ & 0.62 \\
        P-diff
        & $\mstd{9.9}{0.6}$ & $\mstd{0.052}{0.000}$
        & $\mstd{0.404}{0.009}$ & 0.25
        & $\mstd{10.0}{0.1}$ & $\mstd{0.249}{0.000}$
        & $\mstd{0.234}{0.004}$ & 0.29 \\
        DWF
        & $\mstd{10.9}{3.6}$ & $\mstd{0.052}{0.002}$
        & $\mstd{0.531}{0.009}$ & 0.27
        & $\mstd{10.0}{0.1}$ & $\mstd{0.249}{0.001}$
        & $\mstd{0.432}{0.009}$ & 0.37 \\
        $\text{DWF}^{\dagger}$
        & $\mstd{72.9}{6.5}$ & $\mstd{0.145}{0.034}$
        & $\mstd{0.595}{0.015}$ & 0.55
        & $\mstd{39.6}{6.6}$ & $\mstd{0.315}{0.024}$
        & $\mstd{0.671}{0.019}$ & 0.59 \\
        $\text{SANE}_{\text{Gauss}}^{\dagger}$
        & $\mstd{16.9}{5.5}$ & $\mstd{0.054}{0.003}$
        & $\mstd{0.531}{0.018}$ & 0.29
        & $\mstd{10.9}{1.9}$ & $\mstd{0.255}{0.009}$
        & $\mstd{0.482}{0.014}$ & 0.39 \\
        $\text{SANE}_{\text{KDE}}^{\dagger}$
        & $\mstd{93.9}{1.6}$ & $\mstd{0.470}{0.066}$
        & $\mstd{0.941}{0.004}$ & 0.59
        & $\mstd{72.9}{2.5}$ & $\mstd{0.555}{0.054}$
        & $\mstd{0.982}{0.002}$ & 0.64 \\
        \cmidrule(lr){2-4} \cmidrule(lr){5-5}
        \cmidrule(lr){6-8} \cmidrule(lr){9-9}
        \textbf{GMN (Ours)}
        & $\mstd{94.4}{0.4}$ & $\mstd{0.469}{0.022}$
        & $\mstd{0.553}{0.008}$ & \textbf{0.91}
        & $\mstd{75.3}{0.8}$ & $\mstd{0.533}{0.012}$
        & $\mstd{0.630}{0.011}$ & \textbf{0.92} \\
        \bottomrule
    \end{tabular}
    }%
\end{table}

\noindent\textbf{Results.}
Table~\ref{tab:unconditional} reports metrics from
Section~\ref{sec:evaluation}.
MLP, P-diff, and unaligned DWF 
are
near chance;
alignment 
helps
DWF but 
accuracy 
stays
below reference.
SANE's Gaussian mode performs poorly; 
KDE sampling and perturbation retain accuracy despite excessive weight similarity, highlighting why measuring accuracy alone is insufficient.
Our GMN achieves the closest joint match:
JWS of $0.91$/$0.92$, versus
at most
$0.59$/$0.64$
for baselines.

\begin{wraptable}{r}{0.59\textwidth}
    \vspace{-1.5em}
    \caption{The equivariant GMN trained on the same MNIST collection,
    unaligned (numbers from Table~\ref{tab:unconditional}) and aligned
    via Git Re-Basin. Notation as in Table~\ref{tab:unconditional}.}
    \label{tab:alignment}
    \centering
    \vspace{-0.8em}
    \small
    \setlength{\tabcolsep}{3.5pt}
    \begin{tabular}{@{}lcccc@{}}
        \toprule
        \bf Alignment & \bf Acc. (\%) & \bf IoU & \bf WCS & \bf JWS ($\uparrow$) \\
        \midrule
        Unaligned & $\mstd{94.4}{0.4}$ & $\mstd{0.469}{0.022}$ & $\mstd{0.553}{0.008}$ & 0.91 \\
        Aligned & $\mstd{94.3}{0.6}$ & $\mstd{0.464}{0.023}$ & $\mstd{0.552}{0.008}$ & 0.90 \\
        \bottomrule
    \end{tabular}
    \vspace{-1em}
\end{wraptable}

\noindent\textbf{Equivariance and Alignment.}
The objective invariance in Section~\ref{sec:inert} motivates
training without alignment.
Retraining the MNIST GMN on the same collection after
Git Re-Basin alignment yields nearly identical accuracy and similarity
statistics (Table~\ref{tab:alignment}).


\subsubsection{Sample and Model Complexity Analysis}
\label{sec:scaling_law}
Recent studies ask whether scale can substitute for
symmetry~\citep{brehmer2024does}. We investigate this question
in weight-space generation by varying training collection
size and model capacity.

\noindent\textbf{Dataset and setup.}
We vary the training collection from $100$ to $50\mathrm{K}$
independently trained MLPs on
\texttt{mfeat-karhunen}~\citep{bischl2021openml},
comparing our $2$M-parameter GMN with $2$M and $10$M
DWF and DiT models. Both baselines operate directly on
weights without dimensionality reduction.

\begin{figure}[t]
    \centering
    \includegraphics[width=\linewidth]
    {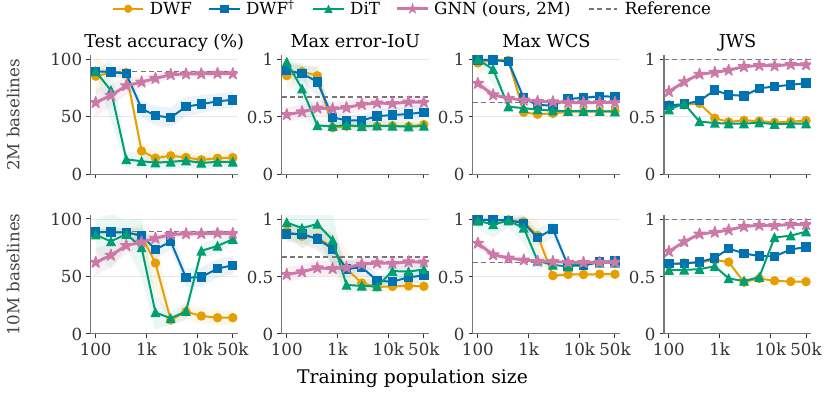}
    \caption{\textbf{Collection and capacity scaling on
    \texttt{mfeat-karhunen}.}
    Baselines have $2$M (top) or $10$M (bottom) parameters.
    Columns report test accuracy, Max error-IoU,
    Max WCS, and JWS.
    Dashed lines 
    are
    collection reference means
    in the first three columns and JWS $=1$ in the last.
    Bands show $\pm1$ standard deviation.
    $\dagger$ denotes post-hoc alignment via Git Re-Basin.
    }
    \label{fig:scaling_law_memorization}
\end{figure}

\noindent\textbf{Results.}
Figure~\ref{fig:scaling_law_memorization} shows that high
baseline accuracy at small collection sizes coincides with
near-unit Max-IoU and WCS, indicating memorization.
Our $2$M GMN achieves the highest JWS throughout the sweep,
approaching the reference performance and similarity
statistics as the collection grows.
Within the evaluated training budgets, increasing baseline
capacity fivefold or applying alignment does not close the
gap in joint generation quality.
Technical details appear in Appendix~\ref{appendix:scaling_complexity}.


\subsection{Conditional Generation}
\label{sec:conditional_generation}
We next test whether a shared velocity field can generate
networks for different tasks and architectures and produce
useful models at domain conditions absent from training.


\subsubsection{Heterogeneous Architecture Generation and unseen width transfer}
\label{sec:heterogeneous_generation}

\noindent\textbf{Datasets and setup.}
Our framework is not limited to networks of the same architecture. We create an
(inter) unalignable
collection
of
$100\mathrm{K}$
independently trained MLPs 
on
$20$ tabular datasets
from OpenML~\citep{bischl2021openml}, with $5\mathrm{K}$
networks per dataset.
The collection spans $16$ 
architectures,
varying in input dimensionality, output classes, depth,
and hidden widths.
Four pairs of tasks share the same architecture, so
architecture alone does not identify the task.
We train a single conditional GNN flow, supplying the
task identity $c$ separately from the parameter graph $\gG$.
All architectures share the velocity network’s parameters, and none are architecture-specific.

\begin{figure}[t]
    \centering
    \includegraphics[width=\textwidth]{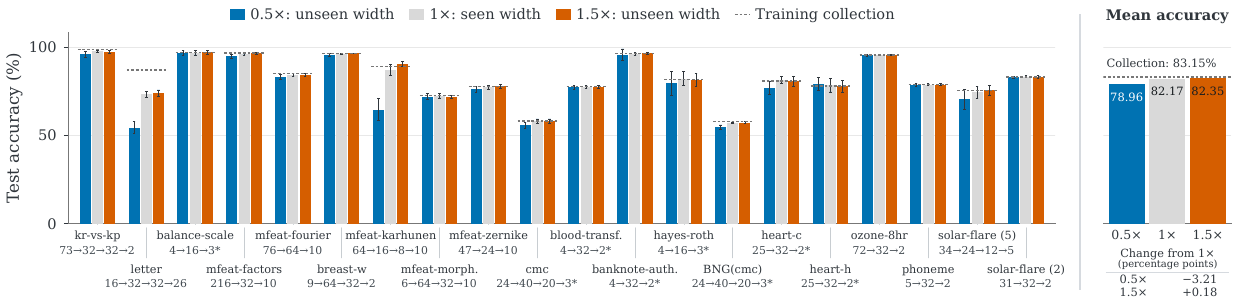}
    \vspace{-2em}
    \caption{\textbf{Heterogeneous generation and zero-shot
    width transfer.}
    A single velocity network with shared parameters
    generates networks across all $20$ tasks.
    Base architectures list layer widths from input to output;
    $*$ marks architectures shared by two tasks.
    Narrow and Wide multiply all hidden widths by
    $0.5\times$ and $1.5\times$, respectively, while
    preserving input and output dimensions.
    Reference reports the mean accuracy of independently
    trained networks at the base architecture.
    Generated accuracies are mean$_{\pm\mathrm{std}}$
    (\%) over $100$ networks per entry.
    The final plot averages the task means.}
    \label{fig:cc18_merged}
\end{figure}

\noindent\textbf{Results.}
Our generated networks achieve
a mean test accuracy of $82.17\%$, compared with the
collection's 
$83.15\%$
(Figure~\ref{fig:cc18_merged}).
Remarkably,
our method can even zero-shot generalize to new architectures it has never seen.
We test this 
by
scaling every hidden width by $0.5\times$ or $1.5\times$.
Only the 
architecture
changes; our velocity 
GMN
requires neither new parameters nor retraining.
We get a
mean accuracy of
$78.96\%$ for narrow and $82.35\%$ for wide networks.
Performance varies across tasks: \texttt{letter} has the
largest gap to its original collection reference, while
width reduction most strongly affects \texttt{letter}
and \texttt{mfeat-karhunen}. Details are provided in Appendix~\ref{appendix:heterogeneous_generation}.

\subsubsection{Domain-Conditioned Generation}
\label{sec:domain_conditioned_generation}

\noindent\textbf{Dataset and setup.}
%
%
Classifiers may 
fail
outside their training domain.
We test 
if
conditional generation yields individual
classifiers 
accurate
across domains.
On
Folktables/ACSIncome~\citep{ding2021retiring} data from
California (CA) and Puerto Rico (PR), 
exhibiting domain shift as Californians tend to be wealthier,
we independently train
$5\mathrm{K}$ classifiers per domain to predict 
if
annual
income exceeds 
\$50K.
We train a conditional flow on their weights only at
$c=0$ (CA) and $c=1$ (PR).
We sample at both endpoints and unseen 
$c\in(0,1)$.
Using ROC AUC on both domains, we compare generated networks
with CA--PR weight averages (with and without Git Re-Basin
alignment) and oracle logit ensembles.
We evaluate JWS against each domain's reference cloud.
Details in
Appendix~\ref{appendix:acs_conditional_generation}. 

\begin{figure}[t]
    \vspace{-0.5em}
    \centering
    \includegraphics[width=0.95\textwidth]{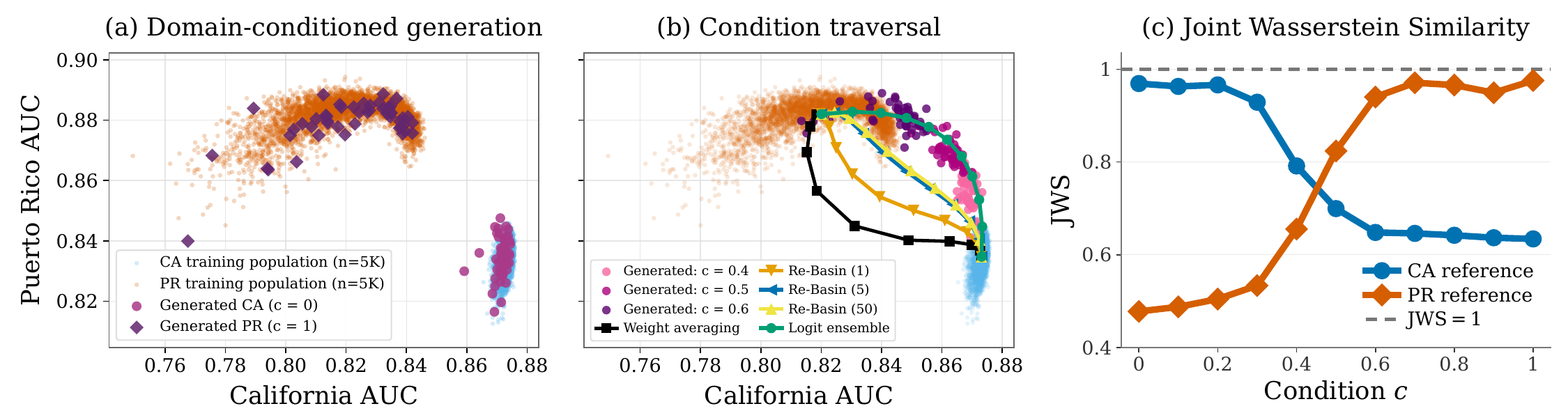}
    \vspace{-0.5em}
    \caption{\textbf{ACS domain conditioning.}
    Faint points show training collections.
    \textbf{(a)} Endpoint generation.
    \textbf{(b)} Intermediate conditions
    $c\in\{0.4,0.5,0.6\}$ and AUC comparison curves.
    Panels (a,b) use $50$ networks per condition,
    with guidance scales $1.5$ and $4.0$, respectively.
    Comparison curves average $50$ CA--PR checkpoint pairs;
    Re-Basin labels denote matching iterations.
    \textbf{(c)} JWS against both domain references,
    using $100$ generated networks at each
    $c\in\{0,0.1,\ldots,1\}$.}
    \label{fig:acs_conditioning}
\end{figure}

\noindent\textbf{Results.}
Endpoint conditioning yields networks with domain-specific performance profiles
(Figure~\ref{fig:acs_conditioning}).
Remarkably, 
at intermediate conditions absent from training, generated networks move along the CA--PR performance trade-off,
achieving cross-domain AUC comparable to two-network logit ensembles.
Unaligned weight averaging degrades performance;
post-hoc alignment improves the mixtures.
As a complementary check, JWS decreases from $0.969/0.976$
at the endpoints to $0.700/0.824$ at $c=0.5$ (CA/PR),
consistent with a shift away from domain reference collections.



\section{Conclusion}
We proposed a permutation-equivariant 
GMN
velocity field for learning weight distributions directly from independently trained networks, without alignment or single-trajectory construction.
Experiments show our accurate generation with joint performance and similarity statistics closer to independently trained networks than 
with any other baseline.
A single conditional model generates weights across heterogeneous architectures, transfers to unseen hidden widths, and 
yields
individual classifiers with cross-domain performance comparable to logit ensembles.
These results support 
incorporating
weight-space permutation symmetries
directly into generative architectures.


\noindent\textbf{Limitations and future work.}
Our experiments focus on
fairly
small 
networks
as a first step to show accurate weight-space generation.
Regardless of scalability, generation should first work
at small scales without memorization.
%
Though we believe results would hold on larger models, scaling 
to them
is left to future work as
dense graph
representations 
cost
substantial memory and 
compute,
potentially 
reduced
with
custom GPU kernels and
optimized sparse operations~\citep{hu2020featgraph}. 
While
graph representations extend to
Transformers~\citep{lim2024graph}, generating foundation-model
weights remains an open challenge. Beyond scaling, conditioning currently uses task identifiers
rather than dataset features, and we do not evaluate generation on unseen tasks. Conditioning on dataset
representations, for example through Set
Transformers~\citep{lee2019set} or
DeepSets~\citep{zaheer2017deep}, remains an open direction. 
The
JWS score jointly
evaluates performance and similarity, 
yet complementary metrics would be welcome to more broadly assess weight space generation.
%
Nevertheless, our work opens exciting avenues to explore in weight space generation.

\bibliographystyle{plainnat}
\bibliography{iclr2027_conference}

\clearpage

\appendix
\startcontents[appendix]

\section*{Appendix Contents}
\printcontents[appendix]{}{1}[2]{}

\section{Theoretical Details}
\label{appendix:proofs}

\subsection{Invariance of the Weight Distribution}
\label{appendix:proof_population}

We restate the weight distribution invariance claim.
Fix an architecture $\gG$ and a training procedure $r$.
Let $T_r:\mathbb{R}^{D_{\gG}}\to\mathbb{R}^{D_{\gG}}$
be a map from an initialization to its final
checkpoint. An initialization distribution $p_{\mathrm{init}}$
induces a weight distribution $p_1$ through
$\vw_1=T_r(\vw_{\mathrm{init}})$, where
$\vw_{\mathrm{init}}\sim p_{\mathrm{init}}$.
Suppose that $p_{\mathrm{init}}$ is $G$-invariant and
$T_r$ is $G$-equivariant:
\begin{equation}
    T_r(\mP_g\vw_{\mathrm{init}})
    = \mP_g T_r(\vw_{\mathrm{init}}),
    \qquad \forall g\in G.
\end{equation}
Then $p_1$ is $G$-invariant.

\begin{proof}
Let $g\in G$ and let
$A\subseteq\mathbb{R}^{D_{\gG}}$ be measurable. 
Then
\begin{align}
    p_1(\mP_g A)
    &= \Pr\!\left(
        T_r(\vw_{\mathrm{init}})\in\mP_g A
    \right) \\
    &= \Pr\!\left(
        \mP_g^{-1}T_r(\vw_{\mathrm{init}})\in A
    \right) \\
    &= \Pr\!\left(
        T_r(\mP_g^{-1}\vw_{\mathrm{init}})\in A
    \right) \\
    &= \Pr\!\left(
        T_r(\vw_{\mathrm{init}})\in A
    \right)
    = p_1(A).
\end{align}
The third equality follows from equivariance of $T_r$,
and the fourth from invariance of $p_{\mathrm{init}}$.
\end{proof}

\begin{remark}[Equivariance of training]
Let $\mP_g$ denote a weight-space symmetry induced by
consistently relabeling hidden units or channels, rather
than an arbitrary permutation of parameter coordinates.
For a differentiable loss invariant under these symmetries,
$f(\mP_g\vw)=f(\vw)$, the chain rule gives
\begin{equation}
    \nabla f(\mP_g\vw)=\mP_g\nabla f(\vw).
\end{equation}
As an example, for the toy loss $f(w_1,w_2)=w_1^2+w_2^2$,
swapping $(1,2)$ to $(2,1)$ swaps the gradient
from $(2,4)$ to $(4,2)$.

Gradient descent therefore commutes with hidden-unit
permutations. This also holds for SGD and Adam when
minibatch losses are permutation-invariant, optimizer
states are initialized at zero, hyperparameters are
shared across interchangeable coordinates, and the
stopping rule is permutation-invariant.
\end{remark}

\begin{remark}[Index-dependent structure]
\label{rem:group_restriction}
The claim applies to any group $G$ under which training
is equivariant. Operations tied to specific neuron indices
may restrict this group. For example, a fixed dropout mask
preserves only permutations that leave its pattern unchanged.
\end{remark}

\begin{remark}[Heterogeneous recipes and conditioning]
\label{rem:mixtures}
A mixture of $G$-invariant weight distributions is also
$G$-invariant. Thus, training recipes may vary across networks,
provided each recipe induces a $G$-invariant distribution.
For conditional generation, the same argument applies
separately to each architecture-task pair $(\gG,c)$,
using the corresponding architecture's permutation group.
\end{remark}

\subsection{Equivariance of the Marginal Velocity Field}
\label{appendix:proof_marginal}


Fix an architecture-task pair $(\gG,c)$.
Let $\vw_0\sim p_0$ and $\vw_1\sim p_1$ be independent,
with $G$-invariant laws and finite second moments.
For $\vw_t=(1-t)\vw_0+t\vw_1$, the population
flow-matching loss at time $t$ is
\begin{equation}
    \mathcal{L}_t(v)
    =
    \E_{\vw_0,\vw_1}
    \left[
        \|v(\vw_t)-(\vw_1-\vw_0)\|_2^2
    \right].
\end{equation}
Its minimizer over vector fields is
the marginal velocity field~\citep{lipman2023flow},
\begin{equation}
    u_t(\vw)=\E[\vw_1-\vw_0\mid\vw_t=\vw].
\end{equation}
We show that $u_t(\mP_g\vw)=\mP_g u_t(\vw)$
for every $g\in G$ and almost every $\vw$.

\begin{proof}
Write $\boldsymbol{\Delta}=\vw_1-\vw_0$.
Independence, invariance, and linearity of interpolation imply
\[
    (\mP_g\vw_t,\mP_g\boldsymbol{\Delta})
    \overset{d}{=}(\vw_t,\boldsymbol{\Delta}).
\]
Thus, using this equality in law and invertibility of $\mP_g$,
\begin{align}
    u_t(\vw)
    &= \E[\mP_g\boldsymbol{\Delta}\mid\mP_g\vw_t=\vw]\\
    &= \mP_g\E[\boldsymbol{\Delta}
        \mid\vw_t=\mP_g^{-1}\vw]\\
    &= \mP_g u_t(\mP_g^{-1}\vw).
\end{align}
Replacing $\vw$ with $\mP_g\vw$ gives the result.
\end{proof}

\subsection{Equivariance of the GMN Velocity Field}
\label{appendix:proof_parameterized}

The construction in Section~\ref{sec:method_gnn} satisfies
\begin{equation}
    v_{\vtheta}(\gG,\mP_g\vw,t,c)
    = \mP_g v_{\vtheta}(\gG,\vw,t,c),
    \qquad g\in G.
\end{equation}

\begin{proof}
We apply the graph-metanetwork equivariance argument
of \citet[Proposition~2 and Appendix~B.6]{lim2024graph};
see also \citet[Sections~2.4--3]{kofinas2024graph}.

Each $g\in G$ preserves the connectivity and structural
roles of $\gG$. Shared encoders and positional encodings
constant on $G$-orbits therefore produce correspondingly
permuted initial node and edge features, while the initial
global feature is unchanged.

Shared message-passing updates with permutation-invariant
aggregation $\bigoplus$ preserve this transformation:
node and edge features are permuted, and global features
remain unchanged. AdaLN also commutes with these
permutations, since it normalizes each feature vector
separately and uses the same conditioning embedding
$\vz(t,c)$ across nodes and edges. Residual connections
preserve equivariance as well.

Consequently, after all $L$ blocks, the shared readout
produces correspondingly permuted parameter velocities,
establishing the claim.
\end{proof}

\subsection{Invariance of the Flow Matching Objective under Alignment}
\label{appendix:proof_inert}

Fix an architecture-task pair $(\gG,c)$.
Suppose $v_{\vtheta}$ is $G$-equivariant and the prior
$p_0$ is $G$-invariant. Under independent sampling
$\vw_0\sim p_0$ and $\vw_1\sim p_1$, replacing each target
by $\mP_{\pi(\vw_1)}\vw_1$, for any measurable map
$\pi:\R^{D_{\gG}}\to G$, leaves the expected
flow-matching objective unchanged.

\begin{proof}
Write the sample loss as
\[
    \ell_t(\va,\vb)
    =
    \left\|
        v_{\vtheta}(\gG,(1-t)\va+t\vb,t,c)
        -(\vb-\va)
    \right\|_2^2.
\]
Fix $t$ and $\vw_1$, and let
$\mP=\mP_{\pi(\vw_1)}$.
Equivariance and orthogonality of $\mP$ imply
\[
    \ell_t(\vw_0,\mP\vw_1)
    = \ell_t(\mP^{-1}\vw_0,\vw_1).
\]
Since $\vw_0$ is independent of $\vw_1$, its distribution
remains $p_0$ when the target is fixed.
Moreover, invariance of $p_0$ means that
$\mP^{-1}\vw_0$ also has distribution $p_0$. Therefore,
\begin{align}
    \E_{\vw_0\sim p_0}
    [\ell_t(\vw_0,\mP\vw_1)]
    &=
    \E_{\vw_0\sim p_0}
    [\ell_t(\mP^{-1}\vw_0,\vw_1)]\\
    &=
    \E_{\vw_0\sim p_0}
    [\ell_t(\vw_0,\vw_1)].
\end{align}
This equality holds for every fixed $t$ and $\vw_1$.
Averaging both sides over time and targets gives
\[
    \E_{t,\vw_0,\vw_1}
    [\ell_t(\vw_0,\mP_{\pi(\vw_1)}\vw_1)]
    =
    \E_{t,\vw_0,\vw_1}
    [\ell_t(\vw_0,\vw_1)],
\]
which is the claimed invariance of the training objective.
\end{proof}


\section{Evaluation Metrics}
\label{appendix:metrics}

Following Section~\ref{sec:evaluation}, we evaluate task
performance, functional similarity, matched weight similarity,
and their joint distribution through JWS.

\subsection{Per-Network Scores}

We consider a fixed task and architecture, with training
collection $\mathcal{T}$. Below, $\mathcal{A}$ denotes the
collection against which a query network is compared.
All functional comparisons use the same test set.
For ACS, evaluation is performed separately for CA and PR,
using the corresponding training collection and test set.

\paragraph{Task performance.}
\label{appendix:metric_acc}
The score $S_{\mathrm{task}}(\vw)$ is test accuracy for
MNIST, CIFAR-10, and OpenML tasks, and positive-class
ROC AUC for ACS.

\paragraph{Max IoU.}
\label{appendix:metric_iou}
Following
\citet{zeng2026generative, gupta2026deepweightflow, wang2024neural},
we measure functional similarity through overlap between
classification errors. Let $\mathcal{E}_{\vw}$ denote the
indices of test examples misclassified by $\vw$. We define
\begin{equation}
    \label{max_iou_def}
    S_{\mathrm{func}}(\vw;\mathcal{A})
    = \mathrm{MaxIoU}(\vw;\mathcal{A})
    = \max_{\vw_j\in\mathcal{A}}
      \frac{|\mathcal{E}_{\vw}\cap\mathcal{E}_{\vw_j}|}
           {|\mathcal{E}_{\vw}\cup\mathcal{E}_{\vw_j}|}.
\end{equation}
Lower values indicate less overlap with the errors of
networks in $\mathcal{A}$. Since this score also depends
on classification error rates, we interpret it jointly
with task performance.

\paragraph{Matched weight cosine similarity.}
\label{appendix:metric_wcs}
We align each $\vw_j\in\mathcal{A}$ to $\vw$ using weight
matching in Git Re-Basin~\citep{ainsworth2022git}.
Writing the resulting permutation as $\mP_{\vw,j}$, we define
\begin{equation}
    \mathrm{WCS}(\vw;\mathcal{A})
    = \max_{\vw_j\in\mathcal{A}}
      \frac{\vw^\top\mP_{\vw,j}\vw_j}
           {\|\vw\|_2\|\mP_{\vw,j}\vw_j\|_2}.
\end{equation}
Values near $1$ indicate similar weight directions after
matching. Since matching is approximate, WCS lower-bounds
the maximum cosine similarity over the comparison collection
and all valid hidden-unit permutations.

\subsection{Training-Collection Statistics}

For a training network $\vw_i\in\mathcal{T}$, we evaluate
task performance directly and compute both similarity
scores against $\mathcal{T}\setminus\{\vw_i\}$.
This leave-one-out comparison excludes trivial self-matches.
The mean task-performance and functional-similarity scores are
\begin{align}
    \overline{S}_{\mathrm{task}}^{\mathrm{train}}
    &= \frac{1}{|\mathcal{T}|}
       \sum_{\vw_i\in\mathcal{T}}
       S_{\mathrm{task}}(\vw_i), \\
    \overline{S}_{\mathrm{func}}^{\mathrm{train}}
    &= \frac{1}{|\mathcal{T}|}
       \sum_{\vw_i\in\mathcal{T}}
       S_{\mathrm{func}}
       (\vw_i;\mathcal{T}\setminus\{\vw_i\}).
\end{align}

Computing WCS for every training network requires a quadratic
number of pairwise comparisons. We therefore estimate its
mean using a uniformly sampled subset
$\mathcal{R}\subset\mathcal{T}$ of $m=1{,}000$ reference
networks:
\begin{equation}
    \widehat{\mu}_{\mathrm{WCS}}^{\mathrm{train}}
    = \frac{1}{m}
      \sum_{\vw_i\in\mathcal{R}}
      \mathrm{WCS}
      (\vw_i;\mathcal{T}\setminus\{\vw_i\}).
\end{equation}
Only the query networks are subsampled; each is compared
against the remaining full training collection.

\subsection{Generated-Network Statistics}

For each generated network $\widetilde{\vw}_k$, we evaluate
task performance directly and compute both similarity
scores against the full training collection $\mathcal{T}$.
For $M$ generated networks, the reported means are
\begin{align}
    \overline{S}_{\mathrm{task}}^{\mathrm{gen}}
    &= \frac{1}{M}\sum_{k=1}^{M}
       S_{\mathrm{task}}(\widetilde{\vw}_k), \\
    \overline{S}_{\mathrm{func}}^{\mathrm{gen}}
    &= \frac{1}{M}\sum_{k=1}^{M}
       S_{\mathrm{func}}(\widetilde{\vw}_k;\mathcal{T}), \\
    \overline{\mathrm{WCS}}^{\mathrm{gen}}
    &= \frac{1}{M}\sum_{k=1}^{M}
       \mathrm{WCS}(\widetilde{\vw}_k;\mathcal{T}).
\end{align}
We also report the standard deviation of each score
across the generated networks.

\subsection{Joint Wasserstein Similarity}
\label{appendix:metric_jws}

Separate metric averages can hide mixtures of memorized
and poorly performing networks. JWS compares the joint
distribution of task performance, functional similarity,
and matched weight similarity.

\paragraph{Reference cloud.}
For a network $\vw$ and comparison collection $\mathcal{A}$,
define
\begin{equation}
    \mathbf{v}(\vw;\mathcal{A})
    =
    \bigl(
        S_{\mathrm{task}}(\vw),
        S_{\mathrm{func}}(\vw;\mathcal{A}),
        \mathrm{WCS}(\vw;\mathcal{A})
    \bigr)^\top.
\end{equation}
The reference cloud uses the uniformly sampled subset
$\mathcal{R}\subset\mathcal{T}$ of $m=1{,}000$ networks
for which all three scores are available.
Each reference network $\vw_j\in\mathcal{R}$ contributes
$\mathbf{v}(\vw_j;\mathcal{T}\setminus\{\vw_j\})$,
whereas each generated network $\widetilde{\vw}_i$
contributes $\mathbf{v}(\widetilde{\vw}_i;\mathcal{T})$.
Thus, only the reference queries are subsampled;
similarity comparisons use the full training collection,
excluding self-comparisons for reference networks.
All three coordinates of each point belong to the same
network. For ACS, we construct a separate reference cloud
for each domain.

\paragraph{Coordinate scaling.}
We compute the coordinate means from these reference
vectors and define
\begin{equation}
    \mu_k
    = \frac{1}{m}
      \sum_{\vw_j\in\mathcal{R}}
      v_k(\vw_j;\mathcal{T}\setminus\{\vw_j\}),
    \qquad
    d_k=\max(\mu_k,1-\mu_k).
\end{equation}
We scale coordinate $k$ by
$\widetilde{v}_k=v_k/d_k$.
The scale $d_k$ is the distance from $\mu_k$ to the farther
endpoint of $[0,1]$, expressing discrepancies relative to
this reference-dependent range.
The same scales apply to generated and reference networks
and remain fixed across reference subsamples and evaluated
methods.

\paragraph{Estimation and normalization.}
As in the main text, let
\begin{equation}
    \mathbf{s}_i
    = \widetilde{\mathbf{v}}
      (\widetilde{\vw}_i;\mathcal{T}),
    \qquad i=1,\ldots,M,
\end{equation}
denote the scaled scores of $M$ generated networks.
For each $b=1,\ldots,B$, we sample $M\leq m$ networks
uniformly without replacement from $\mathcal{R}$.
Writing these networks as
$\{\vw_j^{(b)}\}_{j=1}^{M}$, their scaled scores are
\begin{equation}
    \mathbf{r}_j^{(b)}
    = \widetilde{\mathbf{v}}
      \bigl(\vw_j^{(b)};
      \mathcal{T}\setminus\{\vw_j^{(b)}\}\bigr).
\end{equation}
These scores are selected from the fixed reference cloud;
they are not recomputed against the sampled subset.
We compute
\begin{equation}
    W_2^{(b)}
    =
    \left[
        \min_{g\in S_M}
        \frac{1}{M}\sum_{i=1}^{M}
        \left\|
            \mathbf{s}_i-\mathbf{r}_{g(i)}^{(b)}
        \right\|_2^2
    \right]^{1/2},
\end{equation}
where $S_M$ is the permutation group on $M$ indices.
We solve this one-to-one matching using exact linear
assignment with squared-Euclidean costs and report
\begin{equation}
    \mathrm{JWS}
    =
    \max\left\{
        0,\,
        1-\frac{1}{B\sqrt{3}}
        \sum_{b=1}^{B}W_2^{(b)}
    \right\}.
\end{equation}
Division by $B$ averages over reference subsamples, while
division by $\sqrt{3}$ expresses each distance as a
root-mean-square discrepancy per coordinate.
Clipping keeps JWS in $[0,1]$.
Higher scores indicate closer agreement in the joint
performance and similarity statistics.
Across experiments, we use $B=100$ reference subsamples
per JWS estimate, with equal generated sample sizes
across methods within each experiment.

\subsection{Validation of JWS}
\label{appendix:jws_validation}

\paragraph{Sanity check.}
We test whether JWS assigns high scores to held-out networks
trained using the same procedure as the original collections.
We train $100$ additional MNIST MLPs, $100$ CIFAR-10 CNNs,
and $1{,}000$ \texttt{mfeat-karhunen} MLPs using the original
architectures and recipes with new seeds, following
Table~\ref{tab:unconditional_collections} and
Section~\ref{appendix:scaling_complexity_collections}.
These networks are excluded from generator training and
reference construction.
We evaluate them against the original training collections,
keeping the reference clouds and normalization scales fixed.
Their high JWS scores (Table~\ref{tab:jws_validation}(a))
provide a positive control: independently trained, held-out
networks closely reproduce the reference joint statistics.

\paragraph{Normalization ablation.}
Table~\ref{tab:jws_validation} (b)
ablates coordinate scaling
and the $\sqrt{3}$ distance normalizer.
Without coordinate scaling, the gap between GMN and the
strongest baseline decreases from $0.32$ to $0.18$ on
MNIST and from $0.28$ to $0.17$ on CIFAR-10.
Removing $\sqrt{3}$ clips five methods per dataset to zero,
obscuring differences between them.
GMN ranks highest under all three variants, showing that
its advantage persists across these normalization choices.

\begin{table}
    \centering
    \caption{\textbf{JWS validation.}
    \textbf{(a)} Held-out networks use the original architectures
    and training recipes with disjoint seeds, evaluated using
    the original reference clouds and scales.
    $M$ is the number of held-out networks evaluated.
    \textbf{(b)} Full uses both normalizations; Raw removes
    coordinate scaling; No $\sqrt{3}$ removes only the final
    distance normalizer. All variants retain clipping at zero.
    $\dagger$ denotes aligned training weights.
    Bold marks the highest score among evaluated methods.}
    \label{tab:jws_validation}
    \small
    \setlength{\tabcolsep}{5pt}

    \begin{tabular*}{\linewidth}
        {@{\extracolsep{\fill}}lccccc@{}}
        \multicolumn{6}{@{}l}{
            \textbf{(a) Held-out network sanity check}
        } \\[3pt]
        \toprule
        \textbf{Dataset}
        & \textbf{MNIST}
        & \textbf{CIFAR-10}
        & \multicolumn{3}{c}{\texttt{mfeat-karhunen}} \\
        \cmidrule(lr){4-6}
        $\boldsymbol{M}$
        & $100$ & $100$ & $100$ & $500$ & $1{,}000$ \\
        \midrule
        \textbf{JWS} $\uparrow$
         & $0.990$ & $0.995$ & $0.981$ & $0.989$ & $0.991$ \\
        \bottomrule
    \end{tabular*}

    \par\vspace{0.8em}

    \begin{tabular*}{\linewidth}
        {@{\extracolsep{\fill}}lcccccc@{}}
        \multicolumn{7}{@{}l}{
            \textbf{(b) Normalization ablation}
        } \\[3pt]
        \toprule
        & \multicolumn{3}{c}{\textbf{MNIST}}
        & \multicolumn{3}{c}{\textbf{CIFAR-10}} \\
        \cmidrule(lr){2-4} \cmidrule(lr){5-7}
        \textbf{Method}
        & \textbf{Full}
        & \textbf{Raw}
        & \textbf{No $\sqrt{3}$}
        & \textbf{Full}
        & \textbf{Raw}
        & \textbf{No $\sqrt{3}$} \\
        \midrule
        Random init.
        & $0.15$ & $0.38$ & $0.00$
        & $0.31$ & $0.51$ & $0.00$ \\
        Perturbation
        & $0.54$ & $0.74$ & $0.20$
        & $0.63$ & $0.77$ & $0.36$ \\
        MLP
        & $0.23$ & $0.41$ & $0.00$
        & $0.25$ & $0.48$ & $0.00$ \\
        DiT
        & $0.54$ & $0.72$ & $0.20$
        & $0.62$ & $0.73$ & $0.34$ \\
        P-diff
        & $0.25$ & $0.42$ & $0.00$
        & $0.29$ & $0.50$ & $0.00$ \\
        DWF
        & $0.27$ & $0.42$ & $0.00$
        & $0.37$ & $0.54$ & $0.00$ \\
        DWF$^\dagger$
        & $0.55$ & $0.72$ & $0.22$
        & $0.59$ & $0.71$ & $0.29$ \\
        SANE Gauss$^\dagger$
        & $0.29$ & $0.46$ & $0.00$
        & $0.39$ & $0.55$ & $0.00$ \\
        SANE KDE$^\dagger$
        & $0.59$ & $0.77$ & $0.28$
        & $0.64$ & $0.78$ & $0.37$ \\
        \midrule
        \textbf{GMN (ours)}
        & $\mathbf{0.91}$ & $\mathbf{0.95}$ & $\mathbf{0.84}$
        & $\mathbf{0.92}$ & $\mathbf{0.95}$ & $\mathbf{0.86}$ \\
        \bottomrule
    \end{tabular*}
\end{table}

\newpage
\section{Unconditional Generation}
\label{appendix:unconditional_details}

\subsection{Data and Training Collections}
\label{appendix:unconditional_datasets}

We construct collections of independently trained
classifiers using distinct random seeds and retain their
final checkpoints.
Weights and biases use PyTorch's default uniform
initialization~\citep{paszke2019pytorch}.
Table~\ref{tab:unconditional_collections} summarizes the
architectures and training recipes.
Our CNNs use neither batch nor group normalization.
The graph representation can nonetheless encode the affine
parameters $(\gamma,\beta)$ of normalization layers
analogously to bias terms~\citep{lim2024graph}.

\begin{table}[t]
    \caption{Classifier architectures and training settings.
    All affine layers include biases.}
    \label{tab:unconditional_collections}
    \centering
    \small
    \setlength{\tabcolsep}{3.5pt}
    \begin{tabular}{@{}p{0.22\linewidth}
                       p{0.29\linewidth}
                       p{0.40\linewidth}@{}}
        \toprule
        \textbf{Setting} & \textbf{MNIST MLP} & \textbf{CIFAR-10 CNN} \\
        \midrule
        Collection size
        & $10\mathrm{K}$
        & $50\mathrm{K}$ \\

        Architecture
        & MLP $784\to32\to32\to10$, ReLU
        & Three $3\times3$ convolutions,
          channels $(16,32,48)$, padding $1$, ReLU \\

        Pooling and output
        & Linear classifier
        & $2\times2$ max-pooling after the first two
          convolutions; global average pooling;
          linear classifier \\

        Parameters
        & $26{,}506$
        & $19{,}450$ \\

        Optimizer
        & Adam
        & SGD; Nesterov momentum $0.9$;
          weight decay $5\times10^{-4}$ \\

        Learning rate
        & $10^{-3}$, constant
        & $0.1$, cosine decay to zero \\

        Batch size / epochs
        & $64$ / $5$
        & $128$ / $60$ \\

        Loss
        & Cross-entropy
        & Cross-entropy; label smoothing $0.1$ \\

        Input preprocessing
        & Pixels in $[0,1]$
        & Pixels in $[0,1]$; channel normalization \\

        Channel mean
        & --
        & $(0.4914,0.4822,0.4465)$ \\

        Channel std.
        & --
        & $(0.2470,0.2435,0.2616)$ \\

        Training augmentation
        & None
        & Random horizontal flip; $32\times32$ crop
          after four-pixel reflection padding \\
        \bottomrule
    \end{tabular}
\end{table}

\subsection{GMN Configuration}
\label{appendix:unconditional_generation_of_mnist_mlps}
\label{appendix:unconditional_generation_of_cnns}

Both GMNs use the flow-matching objective in
Equation~\ref{eq:training_objective}, with independent
standard Gaussian noise and $t\sim\mathcal{U}[0,1]$.
For these fixed-architecture experiments, we use
input-layer folding on MNIST and vector-valued kernel
edges on CIFAR-10, as described below.
Neither adaptation is used in domain-conditioned
generation (Section~\ref{sec:domain_conditioned_generation})
or heterogeneous-architecture generation
(Section~\ref{sec:heterogeneous_generation}).

\begin{figure}[t]
    \centering
    \includegraphics[width=0.30\textwidth]
        {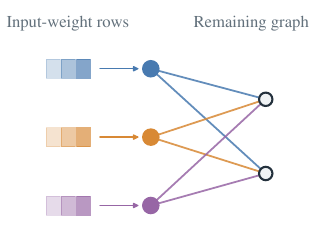}
    \caption{\textbf{Input-layer folding.}
    First-layer weights and biases are encoded as
    first-hidden-layer node features.
    Hidden-unit permutations reorder these features while
    leaving input coordinates fixed; shared encoders and
    decoders therefore preserve permutation equivariance.}
    \label{fig:mnist_input_folding}
\end{figure}

\paragraph{MNIST MLP graph representation.}
Starting from the parameter-graph construction
of~\citet{lim2024graph}, we fold the first affine layer
into node features to reduce message-passing cost
(Figure~\ref{fig:mnist_input_folding}).
For each first-hidden-layer neuron, a shared linear encoder
maps its $784$ incoming weights and bias to a
$128$-dimensional feature. A shared linear decoder maps
its final node feature to the corresponding $785$
parameter velocities.
The remaining $1{,}386$ parameters are represented by
scalar-valued edges on a graph with $74$ neuron nodes
and two bias nodes.
The ODE state retains all $26{,}506$ parameters.

\begin{figure}[t]
    \centering
    \includegraphics[width=0.30\textwidth]
        {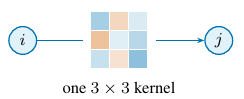}
    \caption{\textbf{CNN kernel edge representation.}
    One edge carries a vectorized $k\times k$ kernel.}
    \label{fig:kernel_inset}
\end{figure}

\paragraph{CIFAR-10 CNN graph representation.}
We build on the parameter-graph construction
of~\citet{lim2024graph}, representing convolutional
channels as nodes.
Following~\citet{kofinas2024graph}, each edge between
an input and output channel carries the nine coefficients
of their $3\times3$ kernel in a fixed spatial order
(Figure~\ref{fig:kernel_inset}).
Shared encoders map these coefficients to edge features,
and shared readouts predict the corresponding nine
parameter velocities.
Fully connected weights and biases are represented by
scalar-valued edges.
Hidden-channel permutations reorder nodes and their
incident edges while leaving the spatial ordering within
each kernel unchanged.

\paragraph{Training and sampling settings.}
Table~\ref{tab:unconditional_gmn} lists the GMN configurations.
All sampling uses the exponential moving average (EMA)
of the learned velocity field.

\begin{table}[t]
    \caption{GMN configurations for unconditional generation.
    Refinement uses corruption time $t^*$ and $k$ cycles.}
    \label{tab:unconditional_gmn}
    \label{tab:mnist_gnn_training}
    \label{tab:cifar_gnn_training}
    \centering
    \small
    \setlength{\tabcolsep}{3.5pt}
    \begin{tabular}{@{}p{0.34\linewidth}
                       p{0.27\linewidth}
                       p{0.30\linewidth}@{}}
        \toprule
        \textbf{Setting} & \textbf{MNIST} & \textbf{CIFAR-10} \\
        \midrule
        \multicolumn{3}{@{}l}{\textbf{Architecture}} \\

        Message-passing blocks
        & $8$
        & $16$ \\

        Node / edge / time dimensions
        & $128$ / $128$ / $128$
        & $192$ / $384$ / $128$ \\

        Positional encodings
        & Orbit-shared
        & Orbit-shared \\

        Weight normalization
        & Parameter-group
        & Parameter-group \\

        Message aggregation
        & Sum
        & Sum \\

        Global aggregation
        & Mean
        & Mean \\

        Parameters (M)
        & $8$
        & $77$ \\

        \midrule
        \multicolumn{3}{@{}l}{\textbf{Optimization}} \\

        Optimizer
        & AdamW
        & AdamW \\

        Learning rate
        & $10^{-3}$
        & $5\times10^{-4}$ \\

        Learning-rate schedule
        & Constant
        & Cosine decay \\

        Warmup updates
        & $0$
        & $2\mathrm{K}$ \\

        Weight decay
        & $0$
        & $0$ \\

        Global batch size
        & $256$
        & $256$ \\

        GPU model
        & NVIDIA RTX A6000
        & NVIDIA L40 \\

        Number of GPUs
        & $4$
        & $8$ \\

        Memory per GPU
        & $48\,\mathrm{GB}$
        & $48\,\mathrm{GB}$ \\

        Training time (wall-clock)
        & $12\,\mathrm{h}$
        & $60\,\mathrm{h}$ \\

        Training updates
        & $200\mathrm{K}$
        & $300\mathrm{K}$ \\

        Gradient clipping
        & Norm $1.0$
        & Norm $1.0$ \\

        EMA decay
        & $0.999$
        & $0.999$ \\

        \midrule
        \multicolumn{3}{@{}l}{\textbf{Sampling}} \\

        Guidance scale~\citep{ho2022classifier}
        & None
        & None \\

        Generated networks
        & $100$
        & $100$ \\

        ODE solver
        & \texttt{dopri5}
        & \texttt{dopri5} \\

        Refinement $(t^*,k)$
        & $(0.75,10)$
        & $(0.3,10)$ \\

        \midrule
        \multicolumn{3}{@{}l}{\textbf{Wall-clock time per network}} \\

        Training from scratch
        & $4\,\mathrm{s}$
        & $86\,\mathrm{s}$ \\

        Generation
        & $0.221\,\mathrm{s}$
        & $3.6\,\mathrm{s}$ \\

        Generation + refinement
        & $0.905\,\mathrm{s}$
        & $25\,\mathrm{s}$ \\
        \addlinespace[2pt]
        \multicolumn{3}{@{}l}{\footnotesize
        Both generation times are amortized over batches of $100$ networks.} \\

        \bottomrule
    \end{tabular}
\end{table}

\subsection{Parameter-Group Normalization}
\label{appendix:parameter_normalization}

We normalize each layer's weights and biases separately.
For each parameter group $r$, we compute a scalar mean $\mu_r$
and standard deviation $\sigma_r$ by pooling all entries across
the training collection, and apply
\[
    \widetilde{\vw}_r
    = \frac{\vw_r-\mu_r\mathbf{1}}{\sigma_r}.
\]
These statistics are permutation-invariant and remain fixed
during flow training and sampling. Generated weights are
transformed back to their original scale before evaluation.

\subsection{Sampling and Refinement}
\label{appendix:unconditional_refinement}
Sampling integrates the EMA velocity field from Gaussian
noise at $t=0$ to $t=1$.
Each refinement cycle~\citep{meng2022sdedit} draws fresh
$\boldsymbol{\epsilon}\sim\mathcal{N}(\vzero,\mI)$,
corrupts the current normalized weights as
\[
    \widetilde{\vw}_{t^*}
    = t^*\vw+(1-t^*)\boldsymbol{\epsilon},
\]
and reintegrates from $t^*$ to $1$ using the same field.
Refinement requires only repeated noise corruption and
integration through the trained flow, with no access to
task data or training networks and no additional training.
We use ten cycles, with $t^*=0.75$ for MNIST and
$t^*=0.3$ for CIFAR-10.
CIFAR-10 refinement uses absolute and relative solver
tolerances of $10^{-5}$.

Table~\ref{tab:gmn_refinement} isolates the effect of
refinement. Even before refinement, our flow achieves
higher JWS, by a large margin, than all evaluated baselines on both datasets
(Table~\ref{tab:unconditional}).
Refinement further improves JWS and increases mean accuracy
by $1.82$ percentage points on MNIST and $3.23$ points
on CIFAR-10.
The refined results are reported in
Table~\ref{tab:unconditional}.

\begin{table}[t]
    \caption{GMN refinement ablation.
    Each setting contains $100$ generated networks;
    MNIST uses the same networks before and after refinement.
    Accuracy, Max IoU, and WCS report
    mean$_{\pm\mathrm{std}}$ across networks;
    JWS uses repeated reference subsampling.}
    \label{tab:gmn_refinement}
    \label{tab:mnist_gnn_refinement}
    \label{tab:cifar_gnn_refinement}
    \label{tab:cifar_gnn_generation}
    \centering
    \small
    \setlength{\tabcolsep}{5pt}
    \begin{tabular}{@{}lcccc@{}}
        \toprule
        \textbf{Sampling}
        & \textbf{Acc. (\%)}
        & \textbf{Max IoU}
        & \textbf{WCS}
        & \textbf{JWS} \\
        \midrule
        \multicolumn{5}{c}{
            \textbf{MNIST}
            \quad $(t^*,k)=(0.75,10)$} \\
        \midrule
        Direct
        & $\mstd{92.58}{0.89}$
        & $\mstd{0.383}{0.029}$
        & $\mstd{0.433}{0.007}$
        & $0.776$ \\
        Refined
        & $\mstd{94.40}{0.41}$
        & $\mstd{0.469}{0.022}$
        & $\mstd{0.553}{0.008}$
        & $0.906$ \\
        \midrule
        \multicolumn{5}{c}{
            \textbf{CIFAR-10}
            \quad $(t^*,k)=(0.3,10)$} \\
        \midrule
        Direct
        & $\mstd{72.11}{1.54}$
        & $\mstd{0.496}{0.019}$
        & $\mstd{0.601}{0.016}$
        & $0.879$ \\
        Refined
        & $\mstd{75.34}{0.80}$
        & $\mstd{0.533}{0.012}$
        & $\mstd{0.630}{0.011}$
        & $0.918$ \\
        \bottomrule
    \end{tabular}
\end{table}


\subsection{Evaluation}
\label{appendix:unconditional_evaluation}
All generated networks are evaluated on the standard test sets of MNIST and CIFAR-10, using the metrics and reference conventions in
Appendix~\ref{appendix:metrics}.
All WCS evaluations use five Git Re-Basin
iterations per network pair, including training-reference
and generated-network comparisons.
This budget keeps full-collection evaluation tractable
while screening for near-copies of training networks.

\subsection{Diversity of Generated Networks}
\label{appendix:unconditional_diversity}

To check for mode collapse, we measure similarity among $100$
generated networks per dataset. For each network, we compute
its maximum IoU and WCS over the remaining $99$ networks,
excluding self-comparisons.
Table~\ref{tab:unconditional_diversity} shows that these maximum
similarities remain well below one, providing evidence against
collapse to repeated or near-duplicate networks.

\begin{table}[t]
    \caption{Leave-one-out maximum similarity among generated
    networks, reported as mean$_{\pm\mathrm{std}}$.}
    \label{tab:unconditional_diversity}
    \centering
    \small
    \setlength{\tabcolsep}{8pt}
    \begin{tabular}{@{}lcc@{}}
        \toprule
        \textbf{Dataset}
        & \textbf{Max IoU}
        & \textbf{WCS} \\
        \midrule
        MNIST
        & $\mstd{0.418}{0.016}$
        & $\mstd{0.529}{0.010}$ \\
        CIFAR-10
        & $\mstd{0.506}{0.009}$
        & $\mstd{0.610}{0.014}$ \\
        \bottomrule
    \end{tabular}
\end{table}

\subsection{Baseline Configurations and Sweeps}
\label{appendix:baselines}


Our baselines cover five families of weight-space generators.
MLP and DiT provide non-equivariant architecture baselines,
DWF performs flow matching directly on weights,
and SANE and P-diff generate in latent space using
fitted priors and learned diffusion, respectively.
P-diff also represents the trajectory-based approach to
a priori alignment, while SANE and DWF
cover post-hoc alignment and permutation augmentation.

All baselines use the same independently trained MNIST and
CIFAR-10 collections and evaluation protocol.
Post-hoc alignment is applied only to designated baseline
configurations; our method always uses unaligned collections except in the alignment ablation (Table~\ref{tab:alignment}).
Before running the sweeps, we reproduce the baselines in their
original settings.

Across these five families, we evaluate $110$ configurations:
$81$ DWF, $18$ SANE, $7$ DiT, and one MLP and P-diff
configuration per dataset.
The searches vary model capacity, learning rate, batch size,
schedule, EMA decay, alignment, and generation prior
where applicable.
Configuration counts include alternative sampling priors:
SANE evaluates each of nine trained autoencoders with two priors.
Table~\ref{tab:baseline_search_coverage} summarizes the search
coverage; Table~\ref{tab:baseline_sweep_results} reports the
outcomes and identifies the configurations selected for
Table~\ref{tab:unconditional}.
The sweeps include baseline models with parameter counts
below, comparable to, and above those of our GMN.
We also evaluate random initialization and perturbed
training networks as controls.

\begin{table}[!htbp]
    \caption{Baseline search coverage. Counts denote evaluated
    configurations, including generation-prior choices.
    The MNIST DWF search is a full Cartesian grid;
    the exact DiT and CIFAR-10 DWF combinations appear in
    Table~\ref{tab:baseline_sweep_results}.}
    \label{tab:baseline_search_coverage}
    \centering
    \small
    \setlength{\tabcolsep}{3pt}
    \renewcommand{\arraystretch}{1.12}
    \begin{tabular}{@{}lcc>{\raggedright\arraybackslash}p{0.60\linewidth}@{}}
        \toprule
        \textbf{Baseline} & \textbf{MNIST} & \textbf{CIFAR-10}
        & \textbf{Search coverage} \\
        \midrule
        MLP & $1$ & $1$
        & Dataset-specific fixed architectures with $21.4$M and
          $80.4$M parameters, respectively. \\
        \addlinespace[3pt]
        DiT & $5$ & $2$
        & MNIST: widths $\{256,368\}$, learning rates
          $\{10^{-3},5\times10^{-4},3\times10^{-4}\}$,
          constant/cosine schedules, and EMA decays
          $\{0.999,0.9999\}$ in five tested combinations.
          CIFAR-10: two width-$864$ configurations. \\
        \addlinespace[3pt]
        DWF & $72$ & $9$
        & MNIST: full grid of $3$ batch sizes, $3$ learning rates,
          $4$ widths, and $2$ alignment settings.
          CIFAR-10: $9$ combinations of $3$ learning rates,
          $2$ batch sizes, and $2$ alignment settings. \\
        \addlinespace[3pt]
        SANE & $14$ & $4$
        & Seven MNIST widths ($2.0$--$202.7$M parameters) and
          two CIFAR-10 widths ($79.7$--$202.7$M), each evaluated
          with KDE and Gaussian priors. \\
        \addlinespace[3pt]
        P-diff & $1$ & $1$
        & Dataset-specific VAE/DDPM configurations, with
          $23.1$M and $42.3$M total parameters, respectively. \\
        \midrule
        \textbf{Total} & $\mathbf{93}$ & $\mathbf{17}$
        & $\mathbf{110}$ evaluated configurations. \\
        \bottomrule
    \end{tabular}
\end{table}

\noindent\textbf{Random initialization and perturbations.}
As controls, we sample random weights and perturb training networks
(Table~\ref{tab:random_init_perturbation}).
We select the perturbation level with the highest JWS:
$\sigma=0.25$ for MNIST and $\sigma=0.10$ for CIFAR-10.

\begin{table}[t]
    \caption{Random initialization and perturbation controls.
    Entries report mean $\pm$ std across networks, or reference
    subsamples for JWS. Bold marks the highest JWS per dataset.}
    \label{tab:random_init_perturbation}
    \centering
    \small
    \setlength{\tabcolsep}{4pt}
    \begin{tabular}{@{}lcccc@{}}
        \toprule
        \textbf{Control} & \textbf{Acc. (\%)} & \textbf{Max IoU}
        & \textbf{WCS} & \textbf{JWS} \\
        \midrule
        \multicolumn{5}{@{}l}{\textbf{MNIST}} \\
        Random init. & $\mstd{10.11}{2.39}$ & $\mstd{0.052}{0.002}$ & $\mstd{0.138}{0.003}$ & $\mstd{0.1457}{0.0010}$ \\
        $\sigma=0.05$ & $\mstd{95.81}{0.25}$ & $\mstd{0.855}{0.029}$ & $\mstd{0.999}{0.000}$ & $\mstd{0.4482}{0.0011}$ \\
        $\sigma=0.10$ & $\mstd{95.52}{0.35}$ & $\mstd{0.717}{0.047}$ & $\mstd{0.995}{0.000}$ & $\mstd{0.5155}{0.0010}$ \\
        $\sigma=0.25$ & $\mstd{92.17}{2.15}$ & $\mstd{0.403}{0.073}$ & $\mstd{0.971}{0.001}$ & $\mstd{\mathbf{0.5366}}{0.0010}$ \\
        $\sigma=0.50$ & $\mstd{76.24}{7.42}$ & $\mstd{0.166}{0.044}$ & $\mstd{0.896}{0.003}$ & $\mstd{0.4471}{0.0012}$ \\
        \midrule
        \multicolumn{5}{@{}l}{\textbf{CIFAR-10}} \\
        Random init. & $\mstd{10.09}{1.71}$ & $\mstd{0.254}{0.012}$ & $\mstd{0.255}{0.006}$ & $\mstd{0.3071}{0.0008}$ \\
        $\sigma=0.05$ & $\mstd{78.29}{0.54}$ & $\mstd{0.767}{0.031}$ & $\mstd{0.999}{0.000}$ & $\mstd{0.5998}{0.0012}$ \\
        $\sigma=0.10$ & $\mstd{74.84}{1.72}$ & $\mstd{0.610}{0.047}$ & $\mstd{0.995}{0.000}$ & $\mstd{\mathbf{0.6296}}{0.0012}$ \\
        $\sigma=0.25$ & $\mstd{53.66}{6.59}$ & $\mstd{0.362}{0.029}$ & $\mstd{0.971}{0.001}$ & $\mstd{0.5413}{0.0008}$ \\
        $\sigma=0.50$ & $\mstd{24.10}{5.03}$ & $\mstd{0.278}{0.015}$ & $\mstd{0.896}{0.002}$ & $\mstd{0.4204}{0.0005}$ \\
        \bottomrule
    \end{tabular}
\end{table}

\paragraph{MLP.}\label{appendix:baseline_mlp}
We use an MLP velocity network on vectorized, unaligned weights,
with AdaLN time conditioning and no built-in permutation equivariance.
The MNIST and CIFAR-10 models use hidden widths $352/1184$,
$8/10$ blocks, and time-embedding dimensions $128/256$,
respectively. Their parameter counts are $21{,}434{,}730$
and $80{,}356{,}378$.
Both use AdamW with learning rate $2\times10^{-4}$, batch size $64$,
EMA decay $0.999$, and parameter-group normalization. Each configuration generates $100$ networks.

\paragraph{DiT.}\label{appendix:baseline_dit}
The DiT velocity network divides vectorized weights into fixed-size
chunks and uses learned positional embeddings and AdaLN time conditioning.
It operates on unaligned weights without built-in permutation equivariance.
All configurations use $8$ transformer blocks and batch size $256$.
Widths $256$, $368$, and $864$ give $8{,}223{,}008$,
$15{,}762{,}848$, and $77{,}913{,}152$ parameters, respectively.
Training budgets are $200$K updates for MNIST and $300$K for
CIFAR-10; checkpoints are selected according to generation quality. We use the official implementation by~\citet{peebles2023scalable}. The configuration that achieved the highest mean JWS determines the entry in Table~\ref{tab:unconditional}. We note that the best run for DiT on MNIST classifiers has two times the amount of parameters as our GMN (Table ~\ref{tab:unconditional_gmn}), and still didn't close the gap. All runs use the AdamW optimizer without weight decay (in line with~\citet{peebles2023scalable}). Each configuration generates $100$ networks.

\paragraph{DWF.}\label{appendix:baseline_dwf}
We use the authors' implementation~\citep{gupta2026deepweightflow}
directly on vectorized weights without PCA, with and without
Git Re-Basin alignment. The MNIST grid uses
$b\in\{32,64,256\}$,
$\eta\in\{10^{-3},5\times10^{-4},10^{-4}\}$, and widths
$h\in\{200,300,400,558\}$ in both alignment settings,
with $200$K updates per run. The widths $200,300,400,558$ give $10{,}687{,}494$, $16{,}045{,}844$, $21{,}424{,}194$, and $29{,}962{,}751$ parameters, respectively.
CIFAR-10 uses width $1954$ ($79{,}992{,}165$ parameters)
and $300$K updates per run. At batch size $256$, all three
learning rates are tested in both alignment settings;
at batch size $64$, $\eta=5\times10^{-4}$ is tested in both
settings and $\eta=10^{-4}$ only with alignment.
Remaining settings follow the authors' implementation. At each configuration we evaluate $100$ generated networks.

\paragraph{SANE.}\label{appendix:baseline_sane}
We use the authors' implementation~\citep{schurholt2024sane}
with permutation augmentation on Git Re-Basin-aligned collections.
MNIST models use $6$ transformer layers and widths
$\{104,160,224,232,320,328,1024\}$; CIFAR-10 models use
$8$ layers and widths $\{640,1024\}$.
Each autoencoder is evaluated with KDE and Gaussian priors,
generating $100$ networks per configuration.
For each dataset and prior, the highest mean JWS determines
the entry in Table~\ref{tab:unconditional}.
Remaining settings follow the authors' implementation.

\paragraph{P-diff.}\label{appendix:baseline_pdiff}
We use the authors' implementation~\citep{wang2024neural}
on unaligned collections. The MNIST model has $23.1$M
parameters ($15.4$M VAE and $7.7$M DDPM), while the
CIFAR-10 model has $42.3$M ($32.7$M VAE and $9.6$M DDPM).
MNIST follows the authors' recipe with $2$K VAE and $8$K DDPM
updates at batch size $50$. CIFAR-10 uses $20$K VAE and $300$K
DDPM updates at batch size $256$, matching the DDPM budget and
batch size to our GMN. Each configuration generates $100$ networks.


\begingroup
\small
\setlength{\tabcolsep}{3pt}
\setlength{\LTcapwidth}{0.95\linewidth}
\setlength{\LTleft}{\fill}
\setlength{\LTright}{\fill}
\renewcommand{\arraystretch}{1.08}
\begin{longtable}{@{}>{\raggedright\arraybackslash}p{0.27\linewidth}ccccc@{}}
\caption{Consolidated baseline configurations and sweep results.
$h$ denotes hidden or transformer width, $\eta$ learning rate, $\rho$ EMA decay. \(b\) batch size, \(L\) the number of blocks, and \(d_t\) the time-embedding dimension. Parameter counts are approximate and given
in millions. Entries use $\mstd{x}{s}\equiv x\pm s$;
accuracy, Max IoU, and WCS summarize generated networks.
JWS is reported as a mean with subsampling standard deviation. Bold marks selected
configurations reported in Table~\ref{tab:unconditional}.
The $36$ unaligned MNIST DWF runs are summarized by per-metric
extrema, which need not correspond to the same configuration.}
\label{tab:baseline_sweep_results}
\label{tab:runs_mlp}\label{tab:runs_dit}
\label{tab:runs_dwf}\label{tab:runs_dwf_cifar}
\label{tab:runs_sane}\label{tab:runs_pdiff}\\
\toprule
\textbf{Configuration} & \textbf{Params (M)} & \textbf{Acc. (\%)}
& \textbf{Max IoU} & \textbf{WCS} & \textbf{JWS} \\
\midrule
\endfirsthead
\multicolumn{6}{@{}l}{\textbf{Table \thetable{} (continued)}}\\
\toprule
\textbf{Configuration} & \textbf{Params (M)} & \textbf{Acc. (\%)}
& \textbf{Max IoU} & \textbf{WCS} & \textbf{JWS} \\
\midrule
\endhead
\midrule
\multicolumn{6}{r@{}}{\footnotesize Continued on next page}\\
\endfoot
\bottomrule
\endlastfoot
\addlinespace[4pt]
\multicolumn{6}{@{}l}{\textbf{MLP: MNIST}}\\*
\addlinespace[2pt]
$h=352,\ L=8$\newline $d_t=128$ & $21.4$ & $\mstd{9.78}{3.74}$ & $\mstd{0.0517}{0.0020}$ & $\mstd{0.3554}{0.0086}$ & $\boldsymbol{\mstd{0.2335}{0.0010}}$ \\
\addlinespace[4pt]
\multicolumn{6}{@{}l}{\textbf{MLP: CIFAR-10}}\\*
\addlinespace[2pt]
$h=1184,\ L=10$\newline $d_t=256$ & $80.4$ & $\mstd{9.98}{1.55}$ & $\mstd{0.2558}{0.0120}$ & $\mstd{0.1464}{0.0025}$ & $\boldsymbol{\mstd{0.2520}{0.0009}}$ \\
\addlinespace[4pt]
\multicolumn{6}{@{}l}{\textbf{DiT: MNIST}}\\*
\addlinespace[2pt]
$h=256,\ \eta=10^{-3}$\newline Cosine; $\rho=0.999$ & $8.2$ & $\mstd{9.95}{5.40}$ & $\mstd{0.0517}{0.026}$ & $\mstd{0.5345}{0.009}$ & $\mstd{0.2625}{0.0010}$ \\
$h=256,\ \eta=5 \times 10^{-4}$\newline Cosine; $\rho=0.999$ & $8.2$ & $\mstd{9.10}{4.73}$ & $\mstd{0.0513}{0.024}$ & $\mstd{0.5324}{0.009}$ & $\mstd{0.2587}{0.0010}$ \\
$h=368,\ \eta=10^{-3}$\newline Cosine; $\rho=0.999$ & $15.8$ & $\mstd{71.97}{6.45}$ & $\mstd{0.1381}{0.032}$ & $\mstd{0.5553}{0.010}$ & $\boldsymbol{\mstd{0.5401}{0.0013}}$ \\
$h=368,\ \eta=5 \times 10^{-4}$\newline Cosine; $\rho=0.999$ & $15.8$ & $\mstd{62.78}{8.93}$ & $\mstd{0.1086}{0.024}$ & $\mstd{0.5509}{0.009}$ & $\mstd{0.4910}{0.0013}$ \\
$h=368,\ \eta=3 \times 10^{-4}$\newline Const.; $\rho=0.9999$ & $15.8$ & $\mstd{66.06}{7.49}$ & $\mstd{0.1179}{0.024}$ & $\mstd{0.5515}{0.009}$ & $\mstd{0.5081}{0.0013}$ \\
\addlinespace[4pt]
\multicolumn{6}{@{}l}{\textbf{DiT: CIFAR-10}}\\*
\addlinespace[2pt]
$h=864,\ \eta=2 \times 10^{-4}$\newline Const.; $\rho=0.999$ & $77.9$ & $\mstd{32.47}{5.98}$ & $\mstd{0.2952}{0.019}$ & $\mstd{0.5377}{0.015}$ & $\mstd{0.5400}{0.0006}$ \\
$h=864,\ \eta=5 \times 10^{-4}$\newline Cosine.; $\rho=0.999$ & $77.9$ & $\mstd{43.67}{5.52}$ & $\mstd{0.326}{0.025}$ & $\mstd{0.553}{0.015}$ & $\boldsymbol{\mstd{0.620}{0.0006}}$ \\
\addlinespace[4pt]
\multicolumn{6}{@{}l}{\textbf{DWF: MNIST, aligned, $b=32$}}\\*
\addlinespace[2pt]
$h=200,\ \eta=10^{-3}$ & $10.7$ & $\mstd{72.78}{6.67}$ & $\mstd{0.1452}{0.033}$ & $\mstd{0.6270}{0.028}$ & $\mstd{0.5421}{0.0013}$ \\
$h=300,\ \eta=10^{-3}$ & $16.0$ & $\mstd{71.82}{8.29}$ & $\mstd{0.1419}{0.035}$ & $\mstd{0.5936}{0.025}$ & $\mstd{0.5403}{0.0013}$ \\
$h=400,\ \eta=10^{-3}$ & $21.4$ & $\mstd{72.11}{7.17}$ & $\mstd{0.1416}{0.032}$ & $\mstd{0.5848}{0.012}$ & $\mstd{0.5425}{0.0013}$ \\
$h=558,\ \eta=10^{-3}$ & $30.0$ & $\mstd{65.18}{12.32}$ & $\mstd{0.1190}{0.031}$ & $\mstd{0.5563}{0.044}$ & $\mstd{0.5020}{0.0013}$ \\
$h=200,\ \eta=5 \times 10^{-4}$ & $10.7$ & $\mstd{70.43}{10.66}$ & $\mstd{0.1377}{0.034}$ & $\mstd{0.5955}{0.029}$ & $\mstd{0.5315}{0.0013}$ \\
$h=300,\ \eta=5 \times 10^{-4}$ & $16.0$ & $\mstd{72.85}{6.54}$ & $\mstd{0.1454}{0.034}$ & $\mstd{0.5952}{0.015}$ & $\boldsymbol{\mstd{0.5470}{0.0013}}$ \\
$h=400,\ \eta=5 \times 10^{-4}$ & $21.4$ & $\mstd{69.71}{7.38}$ & $\mstd{0.1317}{0.030}$ & $\mstd{0.5724}{0.015}$ & $\mstd{0.5288}{0.0013}$ \\
$h=558,\ \eta=5 \times 10^{-4}$ & $30.0$ & $\mstd{54.06}{12.37}$ & $\mstd{0.0911}{0.019}$ & $\mstd{0.5217}{0.042}$ & $\mstd{0.4476}{0.0012}$ \\
$h=200,\ \eta=10^{-4}$ & $10.7$ & $\mstd{71.43}{5.93}$ & $\mstd{0.1365}{0.026}$ & $\mstd{0.5947}{0.015}$ & $\mstd{0.5362}{0.0013}$ \\
$h=300,\ \eta=10^{-4}$ & $16.0$ & $\mstd{72.86}{6.72}$ & $\mstd{0.1438}{0.033}$ & $\mstd{0.5861}{0.011}$ & $\mstd{0.5461}{0.0013}$ \\
$h=400,\ \eta=10^{-4}$ & $21.4$ & $\mstd{70.38}{6.68}$ & $\mstd{0.1338}{0.031}$ & $\mstd{0.5728}{0.010}$ & $\mstd{0.5325}{0.0013}$ \\
$h=558,\ \eta=10^{-4}$ & $30.0$ & $\mstd{64.31}{6.83}$ & $\mstd{0.1109}{0.021}$ & $\mstd{0.5542}{0.011}$ & $\mstd{0.4980}{0.0013}$ \\
\addlinespace[4pt]

\multicolumn{6}{@{}l}{\textbf{DWF: MNIST, aligned, $b=64$}}\\*
\addlinespace[2pt]
$h=200,\ \eta=10^{-3}$ & $10.7$ & $\mstd{51.78}{16.84}$ & $\mstd{0.0927}{0.030}$ & $\mstd{0.5239}{0.053}$ & $\mstd{0.4375}{0.0012}$ \\
$h=300,\ \eta=10^{-3}$ & $16.0$ & $\mstd{46.62}{14.84}$ & $\mstd{0.0811}{0.020}$ & $\mstd{0.5152}{0.029}$ & $\mstd{0.4153}{0.0012}$ \\
$h=400,\ \eta=10^{-3}$ & $21.4$ & $\mstd{23.58}{9.19}$ & $\mstd{0.0584}{0.007}$ & $\mstd{0.4543}{0.041}$ & $\mstd{0.3128}{0.0010}$ \\
$h=558,\ \eta=10^{-3}$ & $30.0$ & $\mstd{22.83}{7.55}$ & $\mstd{0.0572}{0.005}$ & $\mstd{0.4645}{0.017}$ & $\mstd{0.3120}{0.0010}$ \\
$h=200,\ \eta=5 \times 10^{-4}$ & $10.7$ & $\mstd{65.01}{9.04}$ & $\mstd{0.1156}{0.032}$ & $\mstd{0.5758}{0.020}$ & $\mstd{0.5019}{0.0013}$ \\
$h=300,\ \eta=5 \times 10^{-4}$ & $16.0$ & $\mstd{61.60}{12.07}$ & $\mstd{0.1099}{0.033}$ & $\mstd{0.5564}{0.022}$ & $\mstd{0.4866}{0.0012}$ \\
$h=400,\ \eta=5 \times 10^{-4}$ & $21.4$ & $\mstd{56.42}{13.54}$ & $\mstd{0.0977}{0.027}$ & $\mstd{0.5402}{0.023}$ & $\mstd{0.4608}{0.0012}$ \\
$h=558,\ \eta=5 \times 10^{-4}$ & $30.0$ & $\mstd{44.30}{13.56}$ & $\mstd{0.0776}{0.019}$ & $\mstd{0.5130}{0.022}$ & $\mstd{0.4062}{0.0012}$ \\
$h=200,\ \eta=10^{-4}$ & $10.7$ & $\mstd{72.35}{7.15}$ & $\mstd{0.1424}{0.032}$ & $\mstd{0.5891}{0.016}$ & $\mstd{0.5434}{0.0013}$ \\
$h=300,\ \eta=10^{-4}$ & $16.0$ & $\mstd{68.36}{7.38}$ & $\mstd{0.1243}{0.029}$ & $\mstd{0.5751}{0.014}$ & $\mstd{0.5188}{0.0013}$ \\
$h=400,\ \eta=10^{-4}$ & $21.4$ & $\mstd{67.45}{8.57}$ & $\mstd{0.1227}{0.029}$ & $\mstd{0.5612}{0.012}$ & $\mstd{0.5149}{0.0013}$ \\
$h=558,\ \eta=10^{-4}$ & $30.0$ & $\mstd{60.68}{9.04}$ & $\mstd{0.1035}{0.023}$ & $\mstd{0.5447}{0.010}$ & $\mstd{0.4809}{0.0012}$ \\
\addlinespace[4pt]

\multicolumn{6}{@{}l}{\textbf{DWF: MNIST, aligned, $b=256$}}\\*
\addlinespace[2pt]
$h=200,\ \eta=10^{-3}$ & $10.7$ & $\mstd{39.64}{13.44}$ & $\mstd{0.0716}{0.018}$ & $\mstd{0.5057}{0.028}$ & $\mstd{0.3853}{0.0011}$ \\
$h=300,\ \eta=10^{-3}$ & $16.0$ & $\mstd{22.40}{8.37}$ & $\mstd{0.0574}{0.006}$ & $\mstd{0.4737}{0.022}$ & $\mstd{0.3113}{0.0010}$ \\
$h=400,\ \eta=10^{-3}$ & $21.4$ & $\mstd{17.78}{6.42}$ & $\mstd{0.0550}{0.004}$ & $\mstd{0.4524}{0.026}$ & $\mstd{0.2889}{0.0010}$ \\
$h=558,\ \eta=10^{-3}$ & $30.0$ & $\mstd{12.31}{4.24}$ & $\mstd{0.0525}{0.002}$ & $\mstd{0.4438}{0.013}$ & $\mstd{0.2643}{0.0010}$ \\
$h=200,\ \eta=5 \times 10^{-4}$ & $10.7$ & $\mstd{56.41}{12.71}$ & $\mstd{0.0967}{0.026}$ & $\mstd{0.5400}{0.026}$ & $\mstd{0.4605}{0.0012}$ \\
$h=300,\ \eta=5 \times 10^{-4}$ & $16.0$ & $\mstd{60.77}{10.48}$ & $\mstd{0.1046}{0.025}$ & $\mstd{0.5466}{0.029}$ & $\mstd{0.4808}{0.0012}$ \\
$h=400,\ \eta=5 \times 10^{-4}$ & $21.4$ & $\mstd{42.34}{14.58}$ & $\mstd{0.0757}{0.018}$ & $\mstd{0.5082}{0.029}$ & $\mstd{0.3969}{0.0011}$ \\
$h=558,\ \eta=5 \times 10^{-4}$ & $30.0$ & $\mstd{28.72}{8.61}$ & $\mstd{0.0614}{0.007}$ & $\mstd{0.4827}{0.018}$ & $\mstd{0.3395}{0.0011}$ \\
$h=200,\ \eta=10^{-4}$ & $10.7$ & $\mstd{68.94}{8.43}$ & $\mstd{0.1267}{0.031}$ & $\mstd{0.5826}{0.016}$ & $\mstd{0.5212}{0.0013}$ \\
$h=300,\ \eta=10^{-4}$ & $16.0$ & $\mstd{62.69}{10.38}$ & $\mstd{0.1107}{0.030}$ & $\mstd{0.5578}{0.015}$ & $\mstd{0.4915}{0.0013}$ \\
$h=400,\ \eta=10^{-4}$ & $21.4$ & $\mstd{60.15}{10.33}$ & $\mstd{0.1029}{0.025}$ & $\mstd{0.5481}{0.014}$ & $\mstd{0.4782}{0.0012}$ \\
$h=558,\ \eta=10^{-4}$ & $30.0$ & $\mstd{53.24}{10.10}$ & $\mstd{0.0877}{0.017}$ & $\mstd{0.5272}{0.014}$ & $\mstd{0.4456}{0.0012}$ \\
\addlinespace[4pt]

\multicolumn{6}{@{}l}{\textbf{DWF: MNIST, unaligned ($36$ configurations; per-metric extrema)}}\\*
\addlinespace[2pt]
Minimum & --- & $9.28$ & $0.051$ & $0.52$ & $\mstd{0.2599}{0.0010}$ \\
Maximum & --- & $11.05$ & $0.052$ & $0.55$ & $\mstd{0.2670}{0.0010}$ \\
\addlinespace[4pt]

\multicolumn{6}{@{}l}{\textbf{DWF: CIFAR-10, aligned, $h=1954$}}\\*
\addlinespace[2pt]
$b=256,\ \eta=5 \times 10^{-4}$ & $80.0$ & $\mstd{27.84}{9.40}$ & $\mstd{0.2891}{0.021}$ & $\mstd{0.6002}{0.030}$ & $\mstd{0.5124}{0.0005}$ \\
$b=256,\ \eta=10^{-3}$ & $80.0$ & $\mstd{32.61}{7.93}$ & $\mstd{0.2952}{0.022}$ & $\mstd{0.6199}{0.022}$ & $\mstd{0.5439}{0.0005}$ \\
$b=256,\ \eta=10^{-4}$ & $80.0$ & $\mstd{38.68}{8.16}$ & $\mstd{0.3125}{0.024}$ & $\mstd{0.6467}{0.023}$ & $\mstd{0.5853}{0.0004}$ \\
$b=64,\ \eta=5 \times 10^{-4}$ & $80.0$ & $\mstd{34.56}{5.62}$ & $\mstd{0.2971}{0.019}$ & $\mstd{0.6422}{0.017}$ & $\mstd{0.5565}{0.0004}$ \\
$b=64,\ \eta=10^{-4}$ & $80.0$ & $\mstd{39.60}{6.62}$ & $\mstd{0.3151}{0.024}$ & $\mstd{0.6713}{0.019}$ & $\boldsymbol{\mstd{0.5905}{0.0004}}$ \\
\addlinespace[4pt]

\multicolumn{6}{@{}l}{\textbf{DWF: CIFAR-10, unaligned, $h=1954$}}\\*
\addlinespace[2pt]
$b=256,\ \eta=5 \times 10^{-4}$ & $80.0$ & $\mstd{10.00}{0.10}$ & $\mstd{0.2487}{0.001}$ & $\mstd{0.4320}{0.009}$ & $\boldsymbol{\mstd{0.3679}{0.0006}}$ \\
$b=256,\ \eta=10^{-3}$ & $80.0$ & $\mstd{9.97}{0.25}$ & $\mstd{0.2487}{0.001}$ & $\mstd{0.4299}{0.007}$ & $\mstd{0.3671}{0.0006}$ \\
$b=256,\ \eta=10^{-4}$ & $80.0$ & $\mstd{10.05}{0.46}$ & $\mstd{0.2488}{0.001}$ & $\mstd{0.4251}{0.007}$ & $\mstd{0.3664}{0.0006}$ \\
$b=64,\ \eta=5 \times 10^{-4}$ & $80.0$ & $\mstd{10.00}{0.02}$ & $\mstd{0.2487}{0.001}$ & $\mstd{0.4252}{0.007}$ & $\mstd{0.3661}{0.0006}$ \\
\addlinespace[4pt]
\multicolumn{6}{@{}l}{\textbf{SANE: MNIST, aligned}}\\*
\addlinespace[2pt]
$h=104$, KDE & $2.0$ & $\mstd{93.86}{1.57}$ & $\mstd{0.4701}{0.066}$ & $\mstd{0.9412}{0.004}$ & $\boldsymbol{\mstd{0.5868}{0.0009}}$ \\
$h=104$, Gauss. & $2.0$ & $\mstd{11.63}{4.35}$ & $\mstd{0.0523}{0.002}$ & $\mstd{0.5344}{0.017}$ & $\mstd{0.2702}{0.0010}$ \\
$h=160$, KDE & $4.3$ & $\mstd{94.99}{0.55}$ & $\mstd{0.5638}{0.054}$ & $\mstd{0.9548}{0.003}$ & $\mstd{0.5847}{0.0009}$ \\
$h=160$, Gauss. & $4.3$ & $\mstd{16.88}{5.45}$ & $\mstd{0.0544}{0.003}$ & $\mstd{0.5310}{0.018}$ & $\boldsymbol{\mstd{0.2931}{0.0010}}$ \\
$h=224$, KDE & $7.9$ & $\mstd{95.15}{0.54}$ & $\mstd{0.5874}{0.057}$ & $\mstd{0.9596}{0.002}$ & $\mstd{0.5785}{0.0009}$ \\
$h=224$, Gauss. & $7.9$ & $\mstd{10.77}{4.27}$ & $\mstd{0.0520}{0.002}$ & $\mstd{0.5789}{0.016}$ & $\mstd{0.2664}{0.0010}$ \\
$h=232$, KDE & $8.5$ & $\mstd{95.17}{0.62}$ & $\mstd{0.5887}{0.057}$ & $\mstd{0.9607}{0.002}$ & $\mstd{0.5772}{0.0009}$ \\
$h=232$, Gauss. & $8.5$ & $\mstd{10.75}{3.78}$ & $\mstd{0.0521}{0.002}$ & $\mstd{0.5630}{0.016}$ & $\mstd{0.2668}{0.0010}$ \\
$h=320$, KDE & $15.7$ & $\mstd{95.58}{0.25}$ & $\mstd{0.6733}{0.042}$ & $\mstd{0.9693}{0.001}$ & $\mstd{0.5548}{0.0009}$ \\
$h=320$, Gauss. & $15.7$ & $\mstd{11.24}{3.59}$ & $\mstd{0.0526}{0.002}$ & $\mstd{0.5902}{0.016}$ & $\mstd{0.2685}{0.0010}$ \\
$h=328$, KDE & $16.4$ & $\mstd{95.44}{0.40}$ & $\mstd{0.6543}{0.055}$ & $\mstd{0.9705}{0.001}$ & $\mstd{0.5574}{0.0009}$ \\
$h=328$, Gauss. & $16.4$ & $\mstd{11.55}{3.49}$ & $\mstd{0.0526}{0.002}$ & $\mstd{0.5909}{0.014}$ & $\mstd{0.2698}{0.0010}$ \\
$h=1024$, KDE & $202.7$ & $\mstd{95.72}{0.27}$ & $\mstd{0.7637}{0.041}$ & $\mstd{0.9754}{0.001}$ & $\mstd{0.5159}{0.0010}$ \\
$h=1024$, Gauss. & $202.7$ & $\mstd{14.30}{3.87}$ & $\mstd{0.0534}{0.002}$ & $\mstd{0.3445}{0.012}$ & $\mstd{0.2497}{0.0010}$ \\
\addlinespace[4pt]
\multicolumn{6}{@{}l}{\textbf{SANE: CIFAR-10, aligned}}\\*
\addlinespace[2pt]
$h=640$, KDE & $79.7$ & $\mstd{72.91}{2.52}$ & $\mstd{0.5546}{0.054}$ & $\mstd{0.9821}{0.002}$ & $\boldsymbol{\mstd{0.6356}{0.0011}}$ \\
$h=640$, Gauss. & $79.7$ & $\mstd{10.92}{1.88}$ & $\mstd{0.2554}{0.009}$ & $\mstd{0.4819}{0.014}$ & $\boldsymbol{\mstd{0.3875}{0.0006}}$ \\
$h=1024$, KDE & $202.7$ & $\mstd{76.00}{1.24}$ & $\mstd{0.6465}{0.047}$ & $\mstd{0.9897}{0.001}$ & $\mstd{0.6338}{0.0012}$ \\
$h=1024$, Gauss. & $202.7$ & $\mstd{11.16}{2.09}$ & $\mstd{0.2565}{0.010}$ & $\mstd{0.4673}{0.015}$ & $\mstd{0.3866}{0.0006}$ \\
\addlinespace[4pt]
\multicolumn{6}{@{}l}{\textbf{P-diff: MNIST}}\\*
\addlinespace[2pt]
VAE/DDPM: $2$K/$8$K\newline $b=50$ & $23.1$ & $\mstd{9.86}{0.63}$ & $\mstd{0.0517}{0.0004}$ & $\mstd{0.4038}{0.0091}$ & $\boldsymbol{\mstd{0.2463}{0.0010}}$ \\
\addlinespace[4pt]
\multicolumn{6}{@{}l}{\textbf{P-diff: CIFAR-10}}\\*
\addlinespace[2pt]
VAE/DDPM: $20$K/$300$K\newline $b=256$ & $42.3$ & $\mstd{10.01}{0.09}$ & $\mstd{0.2487}{0.0003}$ & $\mstd{0.2339}{0.0036}$ & $\boldsymbol{\mstd{0.2944}{0.0008}}$ \\
\end{longtable}
\endgroup


\section{Sample and Model Complexity}
\label{appendix:scaling_complexity}

\paragraph{Data and collections.}
\label{appendix:scaling_complexity_collections}
We independently train $50\mathrm{K}$ MLPs with architecture
$64\to16\to8\to10$, ReLU hidden activations, and $1{,}266$
parameters on \texttt{mfeat-karhunen}~\citep{bischl2021openml}.
Networks differ in training seed and share one stratified
$80/20$ split ($1{,}600$ training and $400$ evaluation examples).
Imputation and standardization are fitted on the training split
and frozen across the collection. Generators use nested checkpoint
subsets with identical indices across methods. Aligned DWF uses
the same indices after Git Re-Basin alignment~\citep{ainsworth2022git}
to a common reference.

\paragraph{Generators.}
We compare a $2$M GMN with DiT and DWF at approximately $2$M
and $10$M parameters, using the same GMN in both comparisons.
Both DWF capacities use aligned and unaligned collections,
yielding seven configurations per collection size and $70$ in total.
We use the authors' DWF implementation~\citep{gupta2026deepweightflow}.
Table~\ref{tab:scaling_hyperparameters} gives all reported
hyperparameters. We use no early stopping or validation-based
checkpoint selection.

\begin{table}[!htbp]
    \caption{Hyperparameters for the scaling experiment.
    Width denotes the GMN feature dimension, DiT hidden dimension,
    or DWF hidden width. Orbit-shared encodings use hidden-unit
    permutation orbits.}
    \label{tab:scaling_hyperparameters}
    \label{tab:scaling_setup}
    \label{tab:scaling_models}
    \centering
    \small
    \setlength{\tabcolsep}{3pt}
    \newcommand{\scalingshared}[1]{\multicolumn{5}{>{\raggedright\arraybackslash}p{\dimexpr0.65\linewidth+24pt\relax}@{}}{#1}}
    \newcommand{\scalingpair}[1]{\multicolumn{2}{>{\centering\arraybackslash}p{\dimexpr0.26\linewidth+6pt\relax}}{#1}}
    \newcommand{\scalingpairlast}[1]{\multicolumn{2}{>{\centering\arraybackslash}p{\dimexpr0.26\linewidth+6pt\relax}@{}}{#1}}
    \begin{tabular}{@{}>{\raggedright\arraybackslash}p{0.27\linewidth}
                        *{5}{>{\centering\arraybackslash}p{0.13\linewidth}}@{}}
        \toprule
        \textbf{Setting} & \textbf{GMN} & \textbf{DiT} & \textbf{DiT-L}
        & \textbf{DWF} & \textbf{DWF-L} \\
        \midrule
        \multicolumn{6}{@{}l}{\textbf{Shared settings}} \\
        Collection sizes $N$
        & \scalingshared{$100$, $200$, $400$, $800$, $1{,}500$,
          $3{,}000$, $6{,}000$, $12{,}000$, $25{,}000$, $50{,}000$} \\
        Training updates & \scalingshared{$75\mathrm{K}$ per configuration} \\
        Global batch size & \scalingshared{$256$} \\
        Learning rate & \scalingshared{$5\times10^{-4}$} \\
        Learning-rate schedule & \scalingshared{Cosine decay to $10^{-6}$} \\
        Gradient clipping & \scalingshared{Global norm $1.0$} \\
        Checkpoint & \scalingshared{Final; EMA weights where specified below} \\
        Generated networks & \scalingshared{$100$ per configuration} \\
        Integration steps & \scalingshared{$100$} \\
        \midrule
        \multicolumn{6}{@{}l}{\textbf{Architecture and representation}} \\
        Parameters & $1{,}958{,}017$ & $2{,}019{,}872$ & $10{,}731{,}424$
                   & $2{,}009{,}538$ & $10{,}557{,}490$ \\
        Feature / hidden width & $64$ & $128$ & $256$ & $608$ & $2{,}176$ \\
        Blocks & $8$ & $8$ & $12$ & -- & -- \\
        Parameter representation & Graph & \scalingpair{$40$ tokens of size $32$}
                                 & \scalingpairlast{Vectorized} \\
        Node / edge dimensions & $64/64$ & -- & -- & -- & -- \\
        Time dimension & $64$ & $64$ & $64$ & $128$ & $128$ \\
        Attention heads & -- & $8$ & $8$ & -- & -- \\
        Message aggregation & Sum & -- & -- & -- & -- \\
        Positional encodings & Orbit-shared & \scalingpair{Learned} & -- & -- \\
        Time conditioning & AdaLN & \scalingpair{AdaLN} & -- & -- \\
        Weight normalization & \multicolumn{3}{c}{Parameter-group} & -- & -- \\
        Collection alignment & \multicolumn{3}{c}{None}
                             & \scalingpairlast{With / without} \\
        Matching iterations & -- & -- & -- & \scalingpairlast{$100$ (aligned)} \\
        Dropout & -- & -- & -- & $0.1$ & $0.1$ \\
        \midrule
        \multicolumn{6}{@{}l}{\textbf{Optimization and sampling}} \\
        Optimizer & AdamW & AdamW & AdamW & AdamW & AdamW \\
        Weight decay & $10^{-4}$ & $10^{-4}$ & $10^{-4}$ & $10^{-5}$ & $10^{-5}$ \\
        EMA decay & $0.999$ & $0.999$ & $0.999$ & None & None \\
        ODE solver & Euler & Euler & Euler & RK4 & RK4 \\
        \bottomrule
    \end{tabular}
\end{table}

\clearpage 
\paragraph{Evaluation.}
All methods use the same held-out examples and the metrics
in Appendix~\ref{appendix:metrics}. Max IoU and matched WCS
use the entire $50\mathrm{K}$ collection at every $N$,
including networks outside the generator's training subset.
JWS uses a fixed cloud of $1{,}000$ reference networks,
excluding self-comparisons from their similarity statistics.
The coordinate scales are fixed at approximately
$(0.890,0.674,0.625)$ for every $N$.
Table~\ref{tab:scaling_all_results} reports all $70$ configurations,
complementing Figure~\ref{fig:scaling_law_memorization}.

\begingroup
\small
\setlength{\tabcolsep}{5pt}
\setlength{\LTleft}{\fill}
\setlength{\LTright}{\fill}
\setlength{\LTcapwidth}{0.95\linewidth}
\captionsetup{font=normalsize}
\begin{longtable}{@{}rlcccc@{}}
\caption{Scaling results for all seven configurations.
Accuracy, Max IoU, and WCS report $\mstd{x}{s}\equiv x\pm s$
across $100$ generated networks; JWS measures joint distributional
agreement with the fixed reference cloud.
Models have approximately $2$M parameters, or $10$M for the -L variants.
$\dagger$ denotes Git Re-Basin alignment of the training collection.
Bold marks the highest JWS at each $N$.}
\label{tab:scaling_all_results}
\label{tab:scaling_full_results}
\label{tab:scaling_dwf_large_results}
\label{tab:scaling_dit_results}
\label{tab:scaling_gnn_results}
\label{tab:scaling_2m}
\label{tab:scaling_10m}\\
\toprule
$N$ & Model & Acc. (\%) & Max IoU & WCS & JWS \\
\midrule
\endfirsthead
\multicolumn{6}{@{}l}{\textbf{Table \thetable{} (continued)}}\\
\toprule
$N$ & Model & Acc. (\%) & Max IoU & WCS & JWS \\
\midrule
\endhead
\midrule
\multicolumn{6}{r@{}}{\footnotesize Continued on next page}\\
\endfoot
\bottomrule
\endlastfoot
$100$ & GMN & $\mstd{62.25}{11.81}$ & $\mstd{0.519}{0.067}$ & $\mstd{0.792}{0.051}$ & $\mathbf{0.721}$ \\*
 & DiT & $\mstd{89.24}{2.47}$ & $\mstd{0.979}{0.025}$ & $\mstd{1.000}{0.000}$ & $0.564$ \\*
 & DiT-L & $\mstd{86.24}{13.33}$ & $\mstd{0.974}{0.101}$ & $\mstd{0.988}{0.069}$ & $0.558$ \\*
 & DWF & $\mstd{85.15}{17.45}$ & $\mstd{0.860}{0.117}$ & $\mstd{0.971}{0.120}$ & $0.605$ \\*
 & DWF$^\dagger$ & $\mstd{89.39}{2.33}$ & $\mstd{0.906}{0.068}$ & $\mstd{0.998}{0.006}$ & $0.600$ \\*
 & DWF-L & $\mstd{88.69}{3.03}$ & $\mstd{0.897}{0.076}$ & $\mstd{0.998}{0.002}$ & $0.603$ \\*
 & DWF-L$^\dagger$ & $\mstd{88.48}{3.11}$ & $\mstd{0.873}{0.070}$ & $\mstd{0.997}{0.006}$ & $0.614$ \\
\addlinespace[3pt]
$200$ & GMN & $\mstd{68.39}{9.73}$ & $\mstd{0.543}{0.062}$ & $\mstd{0.692}{0.032}$ & $\mathbf{0.806}$ \\*
 & DiT & $\mstd{72.68}{25.73}$ & $\mstd{0.745}{0.221}$ & $\mstd{0.916}{0.133}$ & $0.615$ \\*
 & DiT-L & $\mstd{80.46}{24.13}$ & $\mstd{0.924}{0.179}$ & $\mstd{0.958}{0.117}$ & $0.560$ \\*
 & DWF & $\mstd{88.28}{3.42}$ & $\mstd{0.895}{0.062}$ & $\mstd{0.998}{0.002}$ & $0.605$ \\*
 & DWF$^\dagger$ & $\mstd{88.77}{2.44}$ & $\mstd{0.880}{0.066}$ & $\mstd{0.998}{0.002}$ & $0.611$ \\*
 & DWF-L & $\mstd{88.70}{3.87}$ & $\mstd{0.854}{0.095}$ & $\mstd{0.995}{0.020}$ & $0.621$ \\*
 & DWF-L$^\dagger$ & $\mstd{88.55}{3.14}$ & $\mstd{0.874}{0.079}$ & $\mstd{0.997}{0.008}$ & $0.614$ \\
\addlinespace[3pt]
$400$ & GMN & $\mstd{76.82}{7.32}$ & $\mstd{0.577}{0.068}$ & $\mstd{0.663}{0.013}$ & $\mathbf{0.873}$ \\*
 & DiT & $\mstd{12.79}{7.10}$ & $\mstd{0.425}{0.048}$ & $\mstd{0.589}{0.011}$ & $0.459$ \\*
 & DiT-L & $\mstd{86.82}{11.83}$ & $\mstd{0.961}{0.111}$ & $\mstd{0.989}{0.060}$ & $0.566$ \\*
 & DWF & $\mstd{88.27}{7.10}$ & $\mstd{0.859}{0.087}$ & $\mstd{0.994}{0.032}$ & $0.617$ \\*
 & DWF$^\dagger$ & $\mstd{87.21}{5.13}$ & $\mstd{0.806}{0.128}$ & $\mstd{0.983}{0.041}$ & $0.640$ \\*
 & DWF-L & $\mstd{88.38}{3.27}$ & $\mstd{0.843}{0.086}$ & $\mstd{0.997}{0.004}$ & $0.624$ \\*
 & DWF-L$^\dagger$ & $\mstd{88.19}{3.53}$ & $\mstd{0.830}{0.103}$ & $\mstd{0.993}{0.014}$ & $0.630$ \\
\addlinespace[3pt]
$800$ & GMN & $\mstd{79.97}{5.09}$ & $\mstd{0.571}{0.064}$ & $\mstd{0.646}{0.012}$ & $\mathbf{0.888}$ \\*
 & DiT & $\mstd{11.14}{6.97}$ & $\mstd{0.418}{0.039}$ & $\mstd{0.576}{0.012}$ & $0.446$ \\*
 & DiT-L & $\mstd{75.15}{25.56}$ & $\mstd{0.819}{0.229}$ & $\mstd{0.927}{0.136}$ & $0.592$ \\*
 & DWF & $\mstd{20.11}{6.95}$ & $\mstd{0.405}{0.044}$ & $\mstd{0.539}{0.039}$ & $0.489$ \\*
 & DWF$^\dagger$ & $\mstd{56.94}{7.23}$ & $\mstd{0.495}{0.050}$ & $\mstd{0.664}{0.057}$ & $0.733$ \\*
 & DWF-L & $\mstd{85.14}{10.73}$ & $\mstd{0.757}{0.138}$ & $\mstd{0.979}{0.053}$ & $0.646$ \\*
 & DWF-L$^\dagger$ & $\mstd{85.68}{6.43}$ & $\mstd{0.743}{0.132}$ & $\mstd{0.968}{0.063}$ & $0.663$ \\
\addlinespace[3pt]
$1{,}500$ & GMN & $\mstd{83.16}{4.19}$ & $\mstd{0.580}{0.060}$ & $\mstd{0.635}{0.012}$ & $\mathbf{0.908}$ \\*
 & DiT & $\mstd{9.95}{6.13}$ & $\mstd{0.427}{0.037}$ & $\mstd{0.564}{0.013}$ & $0.441$ \\*
 & DiT-L & $\mstd{18.67}{14.96}$ & $\mstd{0.428}{0.073}$ & $\mstd{0.638}{0.073}$ & $0.485$ \\*
 & DWF & $\mstd{13.88}{6.25}$ & $\mstd{0.418}{0.044}$ & $\mstd{0.522}{0.014}$ & $0.456$ \\*
 & DWF$^\dagger$ & $\mstd{50.81}{8.40}$ & $\mstd{0.469}{0.050}$ & $\mstd{0.603}{0.014}$ & $0.692$ \\*
 & DWF-L & $\mstd{61.65}{26.06}$ & $\mstd{0.567}{0.168}$ & $\mstd{0.860}{0.142}$ & $0.628$ \\*
 & DWF-L$^\dagger$ & $\mstd{73.35}{7.98}$ & $\mstd{0.568}{0.073}$ & $\mstd{0.841}{0.073}$ & $0.746$ \\
\addlinespace[3pt]
$3{,}000$ & GMN & $\mstd{86.47}{3.02}$ & $\mstd{0.609}{0.059}$ & $\mstd{0.632}{0.012}$ & $\mathbf{0.939}$ \\*
 & DiT & $\mstd{10.64}{6.38}$ & $\mstd{0.417}{0.039}$ & $\mstd{0.554}{0.013}$ & $0.441$ \\*
 & DiT-L & $\mstd{13.57}{5.80}$ & $\mstd{0.419}{0.040}$ & $\mstd{0.603}{0.012}$ & $0.463$ \\*
 & DWF & $\mstd{15.88}{6.74}$ & $\mstd{0.420}{0.044}$ & $\mstd{0.530}{0.015}$ & $0.469$ \\*
 & DWF$^\dagger$ & $\mstd{49.17}{8.72}$ & $\mstd{0.469}{0.054}$ & $\mstd{0.600}{0.018}$ & $0.683$ \\*
 & DWF-L & $\mstd{12.09}{3.51}$ & $\mstd{0.446}{0.049}$ & $\mstd{0.509}{0.026}$ & $0.452$ \\*
 & DWF-L$^\dagger$ & $\mstd{80.40}{7.21}$ & $\mstd{0.583}{0.074}$ & $\mstd{0.916}{0.077}$ & $0.704$ \\
\addlinespace[3pt]
$6{,}000$ & GMN & $\mstd{87.04}{3.07}$ & $\mstd{0.621}{0.060}$ & $\mstd{0.625}{0.014}$ & $\mathbf{0.949}$ \\*
 & DiT & $\mstd{11.69}{6.26}$ & $\mstd{0.423}{0.041}$ & $\mstd{0.549}{0.012}$ & $0.449$ \\*
 & DiT-L & $\mstd{19.67}{7.67}$ & $\mstd{0.415}{0.043}$ & $\mstd{0.589}{0.016}$ & $0.495$ \\*
 & DWF & $\mstd{14.45}{6.40}$ & $\mstd{0.420}{0.041}$ & $\mstd{0.544}{0.015}$ & $0.463$ \\*
 & DWF$^\dagger$ & $\mstd{58.61}{8.61}$ & $\mstd{0.506}{0.059}$ & $\mstd{0.657}{0.018}$ & $0.750$ \\*
 & DWF-L & $\mstd{19.35}{6.50}$ & $\mstd{0.410}{0.047}$ & $\mstd{0.520}{0.015}$ & $0.484$ \\*
 & DWF-L$^\dagger$ & $\mstd{48.86}{7.86}$ & $\mstd{0.465}{0.050}$ & $\mstd{0.594}{0.014}$ & $0.680$ \\
\addlinespace[3pt]
$12{,}000$ & GMN & $\mstd{87.13}{2.69}$ & $\mstd{0.617}{0.062}$ & $\mstd{0.624}{0.016}$ & $\mathbf{0.946}$ \\*
 & DiT & $\mstd{9.69}{5.94}$ & $\mstd{0.420}{0.033}$ & $\mstd{0.551}{0.013}$ & $0.436$ \\*
 & DiT-L & $\mstd{72.27}{6.91}$ & $\mstd{0.549}{0.060}$ & $\mstd{0.627}{0.012}$ & $0.843$ \\*
 & DWF & $\mstd{12.38}{5.92}$ & $\mstd{0.422}{0.043}$ & $\mstd{0.546}{0.014}$ & $0.452$ \\*
 & DWF$^\dagger$ & $\mstd{60.79}{7.71}$ & $\mstd{0.516}{0.057}$ & $\mstd{0.666}{0.017}$ & $0.765$ \\*
 & DWF-L & $\mstd{15.57}{6.88}$ & $\mstd{0.415}{0.040}$ & $\mstd{0.519}{0.015}$ & $0.464$ \\*
 & DWF-L$^\dagger$ & $\mstd{49.36}{8.61}$ & $\mstd{0.457}{0.041}$ & $\mstd{0.598}{0.014}$ & $0.678$ \\
\addlinespace[3pt]
$25{,}000$ & GMN & $\mstd{87.60}{2.88}$ & $\mstd{0.630}{0.055}$ & $\mstd{0.626}{0.013}$ & $\mathbf{0.958}$ \\*
 & DiT & $\mstd{10.60}{5.59}$ & $\mstd{0.415}{0.041}$ & $\mstd{0.549}{0.011}$ & $0.440$ \\*
 & DiT-L & $\mstd{76.46}{6.49}$ & $\mstd{0.545}{0.060}$ & $\mstd{0.620}{0.014}$ & $0.859$ \\*
 & DWF & $\mstd{13.70}{7.10}$ & $\mstd{0.417}{0.041}$ & $\mstd{0.546}{0.016}$ & $0.458$ \\*
 & DWF$^\dagger$ & $\mstd{63.31}{6.75}$ & $\mstd{0.525}{0.065}$ & $\mstd{0.680}{0.020}$ & $0.780$ \\*
 & DWF-L & $\mstd{14.06}{6.15}$ & $\mstd{0.420}{0.045}$ & $\mstd{0.522}{0.017}$ & $0.458$ \\*
 & DWF-L$^\dagger$ & $\mstd{56.93}{8.09}$ & $\mstd{0.495}{0.059}$ & $\mstd{0.628}{0.014}$ & $0.738$ \\
\addlinespace[3pt]
$50{,}000$ & GMN & $\mstd{87.22}{2.53}$ & $\mstd{0.627}{0.053}$ & $\mstd{0.625}{0.014}$ & $\mathbf{0.955}$ \\*
 & DiT & $\mstd{10.62}{6.15}$ & $\mstd{0.421}{0.038}$ & $\mstd{0.546}{0.013}$ & $0.441$ \\*
 & DiT-L & $\mstd{82.41}{4.50}$ & $\mstd{0.564}{0.061}$ & $\mstd{0.621}{0.013}$ & $0.894$ \\*
 & DWF & $\mstd{14.31}{7.26}$ & $\mstd{0.432}{0.041}$ & $\mstd{0.550}{0.015}$ & $0.467$ \\*
 & DWF$^\dagger$ & $\mstd{64.57}{7.53}$ & $\mstd{0.541}{0.061}$ & $\mstd{0.676}{0.018}$ & $0.795$ \\*
 & DWF-L & $\mstd{14.06}{6.05}$ & $\mstd{0.416}{0.047}$ & $\mstd{0.521}{0.015}$ & $0.456$ \\*
 & DWF-L$^\dagger$ & $\mstd{59.62}{8.07}$ & $\mstd{0.512}{0.056}$ & $\mstd{0.638}{0.013}$ & $0.760$ \\
\end{longtable}
\endgroup

\section{Domain-Conditioned Generation}
\label{appendix:conditional_details}
\label{appendix:acs_conditional_generation}

We study geographic distribution shift on Folktables
ACSIncome~\citep{ding2021retiring}, predicting whether an
individual's annual income exceeds \$50{,}000 in California
and Puerto Rico.
These domains represent different regional populations
under a shared prediction task.
This models a setting where domain-specific classifiers
are available but their original training data cannot be
accessed or pooled.
By learning a conditional distribution over their weights,
we test whether unseen conditioning values can generate
individual networks that balance predictive performance
across both domains.

The flow is trained only on California (CA, $c=0$) and
Puerto Rico (PR, $c=1$) classifiers.
Training collections vary initialization, learning rate,
and training duration, providing a control for mixtures
of training recipes (Remark~\ref{rem:mixtures}).
Table~\ref{tab:acs_setup} summarizes the data, classifier
collections, conditional flow, and AUC evaluation settings.

\begin{table}[t]
    \caption{ACS data, training collections, conditional flow,
    and AUC evaluation settings.}
    \label{tab:acs_setup}
    \label{tab:acs_flow_evaluation}
    \centering
    \small
    \setlength{\tabcolsep}{4pt}
    \begin{tabular}{
        @{}>{\raggedright\arraybackslash}p{0.28\linewidth}
           >{\raggedright\arraybackslash}p{0.68\linewidth}@{}
    }
        \toprule
        \multicolumn{2}{@{}l}{\textbf{Data and training collections}} \\
        Data source
        & Folktables ACSIncome~\citep{ding2021retiring};
          2018 one-year ACS data \\
        Domains and split
        & CA ($c=0$), PR ($c=1$); up to $30\mathrm{K}$ examples
          per domain; stratified $80/20$ training/test split \\
        Categorical features
        & One-hot encode \texttt{COW}, \texttt{MAR}, \texttt{RELP},
          \texttt{SEX}, and \texttt{RAC1P} \\
        Numerical features
        & Median-impute and standardize \texttt{AGEP},
          \texttt{SCHL}, and \texttt{WKHP} \\
        Preprocessing
        & Fitted jointly on both training sets, then frozen;
          $44$ input features \\
        Collection size
        & $5\mathrm{K}$ independently trained networks per domain; default PyTorch
          initialization; final checkpoints only \\
        Classifier architecture
        & MLP $44\to32\to32\to2$; $2{,}562$ parameters \\
        Classifier optimization
        & Adam; cross-entropy loss; batch size $256$ \\
        Learning rate / epochs
        & Per network: log-uniform rate in
          $[3\times10^{-4},3\times10^{-3}]$;
          uniform integer in $[15,40]$ epochs \\
        \midrule
        \multicolumn{2}{@{}l}{\textbf{Conditional flow}} \\
        Training conditions
        & $c\in\{0,1\}$ only \\
        Message-passing blocks
        & $8$ \\
        Parameters (M)
        & Approximately $7.8$ \\
        Node / edge / time dimensions
        & $128$ / $128$ / $128$ \\
        Positional encodings
        & Orbit-shared \\
        Message / global aggregation
        & Attention-weighted / mean \\
        Condition embedding
        & Two-layer SiLU MLP, dimension $128$;
          concatenated with time embedding for per-block AdaLN \\
        Coupling / time sampling
        & Independent Gaussian--target coupling;
          uniform time sampling \\
        Weight normalization
        & Parameter-group \\
        Training updates
        & Approximately $57\mathrm{K}$ \\
        Global batch size
        & $256$ \\
        Learning rate / schedule
        & $2\times10^{-4}$ / cosine decay \\
        EMA decay / condition dropout
        & $0.999$ / $0.1$; EMA weights used for sampling \\
        \midrule
        \multicolumn{2}{@{}l}{\textbf{Generation and AUC evaluation}} \\
        Generated networks
        & $50$ per condition, $c\in\{0,0.4,0.5,0.6,1\}$ \\
        Guidance scale
        & $1.5$ at endpoints; $4.0$ at intermediate conditions \\
        ODE solver / refinement
        & \texttt{dopri5}; no refinement \\
        Evaluation
        & Positive-class ROC AUC
          on both held-out domains \\
        Baseline pairs
        & $50$ CA--PR checkpoint pairs for all baselines \\
        Mixing and reporting
        & $\alpha\in\{0,0.1,\ldots,1\}$;
          curves report mean AUC over the $50$ pairs \\
        Weight averaging
        & Direct or after Git Re-Basin~\citep{ainsworth2022git}
          with $1$, $5$, or $50$ matching iterations \\
        Logit ensembling
        & Average logits before softmax \\
        \bottomrule
    \end{tabular}
\end{table}

\section{Heterogeneous Architecture Generation}
\label{appendix:heterogeneous_generation}

We use one conditional flow to generate networks across $20$ datasets
and $16$ native architectures, and to transfer to unseen hidden-width
configurations. All generations use the same learned flow parameters
and checkpoint, with no padding or architecture-specific learned
parameters.

\subsection{Data and Training Collections}
We use $5\mathrm{K}$ independently trained MLPs per OpenML
dataset~\citep{bischl2021openml}, for $100\mathrm{K}$ networks in total.
The native architectures have one or two hidden layers, input
dimensions from $4$ to $216$, and output dimensions from $2$ to $26$
(Table~\ref{tab:heterogeneous_results}). Each dataset uses a stratified
$80/20$ training--test split. Categorical features are imputed with the
most frequent value and one-hot encoded; numerical features are
median-imputed and standardized. Preprocessing is fitted only on the
training split. Source networks use ReLU after each hidden layer and
retain only their final checkpoints.

\subsection{Shared Representation and Conditioning}
We represent each MLP as a graph following~\citet{lim2024graph}. One permutation-equivariant
GMN parameterizes the velocity field for every architecture: all
encoders, message-passing blocks, and readouts share their learned
parameters, without any architecture specific parameters or padding. Sinusoidal positional encodings use structural roles shared within hidden-unit permutation orbits.

Architecture is specified by the graph, while dataset identity $c$
is linearly encoded and added to the time embedding for AdaLN
conditioning. Dropped conditions use a learned null token.
Four native architectures are each shared by two datasets, so
architecture and task identity are supplied separately.
Table~\ref{tab:heterogeneous_setup} summarizes training and sampling.

\begin{table}[!htbp]
    \caption{Heterogeneous collection construction, shared flow training,
    and evaluation. All datasets and width configurations use the same
    flow checkpoint and sampling settings. All the networks in the training collections were initialized with different seeds.}
    \label{tab:heterogeneous_setup}
    \centering
    \small
    \setlength{\tabcolsep}{4pt}
    \begin{tabular}{@{}>{\raggedright\arraybackslash}p{0.28\linewidth}
                        >{\raggedright\arraybackslash}p{0.68\linewidth}@{}}
        \toprule
        \multicolumn{2}{@{}l}{\textbf{Training collections}} \\
        Collection size & $100\mathrm{K}$ networks; $20$ datasets;
        $16$ native architectures \\
        Classifier optimization & Adam; learning rate $10^{-3}$;
        cross-entropy loss \\
        Training recipe & Batch size / epochs: $128/25$;
        $32/300$ for balance-scale, hayes-roth, and
        solar-flare (OpenML 40686) \\
        \midrule
        \multicolumn{2}{@{}l}{\textbf{Conditional flow}} \\
        Message-passing blocks & $10$ \\
        Parameters (M) & $15.13$ \\
        Node / edge / time dimensions & $128$ / $192$ / $128$ \\
        Activation / edge encoder & Tanh-approximate GELU / Fourier features \\
        Weight normalization & None \\
        Coupling / time sampling & Independent Gaussian--target coupling;
        uniform time sampling \\
        Optimizer & AdamW; learning rate $5\times10^{-4}$;
        weight decay $0$ \\
        Learning-rate schedule & $2\mathrm{K}$-update linear warmup,
        then cosine decay over $1$M scheduled updates \\
        Training updates & $917{,}600$ \\
        Global batch size & $192$ across $6$ GPUs \\
        Gradient clipping & Global norm $1.0$ \\
        EMA decay / condition dropout & $0.999$ / $0.1$ \\
        \midrule
        \multicolumn{2}{@{}l}{\textbf{Evaluation}} \\
        Generated networks & $100$ per dataset and width configuration \\
        Hidden-width multipliers & $0.5$, $1.0$, and $1.5$ \\
        Guidance scale & $1.0$ \\
        ODE solver & \texttt{dopri5}\\
        Scratch references & $5\mathrm{K}$ networks at native widths;
        $100$ at each variant width, using the dataset's source recipe \\
        \bottomrule
    \end{tabular}
\end{table}

\subsection{Generation and Width Transfer}
All networks are sampled using the conditional EMA velocity field
with guidance scale $1.0$.
For zero-shot width transfer, we multiply every hidden width by $0.5$
or $1.5$, preserving depth and input/output dimensions. The resulting
$32$ distinct variant architectures are absent from the flow's training
collection and require only a change to the input graph, without
retraining or adding learned parameters. Dataset identities remain
those seen during flow training.
All results use three refinement cycles with $t^*=0.9$, following
Appendix~\ref{appendix:unconditional_refinement}.

We compare test accuracy with independently trained networks at the
same target architecture, evaluated on the same held-out examples.
Native references contain all $5\mathrm{K}$ source networks; each
variant reference contains $100$ networks trained with the dataset's
source recipe in Table~\ref{tab:heterogeneous_setup}.

\noindent\textbf{Results.}
Table~\ref{tab:heterogeneous_results} reports accuracies and gaps
relative to training from scratch for all $60$ dataset--width configurations.
Averaging equally across datasets, generated networks trail the
corresponding scratch references by $0.98$ percentage points at
native widths and $0.58/1.69$ points at unseen $0.5/1.5\times$ widths.
At unseen $0.5\times$ widths, generation improves over scratch
training on several datasets, including hayes-roth ($+14.03$ points)
and mfeat-karhunen ($+5.12$ points), showing that width transfer can
yield better-performing smaller networks.
Performance varies across tasks, with a $25.16$-point deficit on
letter at this width.
These results demonstrate generalization to unseen widths using
the same flow parameters.

\begin{table}[p]
    \caption{Heterogeneous generation and width transfer.
    Headings give each dataset's native architecture; width multipliers
    apply to all hidden layers, with $0.5\times$ and $1.5\times$ unseen
    during flow training. Accuracy (\%) is mean$_{\pm\mathrm{std}}$;
    $\Delta$ is generated minus scratch accuracy in percentage points.
    All generated networks use one shared flow.
    Scratch references contain $5\mathrm{K}$ native networks or $100$
    networks per variant; each generated entry uses $100$ networks.
    All evaluations use the same held-out split per dataset.
    The final block averages all $20$ datasets equally.
    Solar-flare (5 classes) is OpenML $40686$.}
    \label{tab:heterogeneous_results}
    \label{tab:heterogeneous_accuracy_gap}
    \centering
    \small
    \setlength{\tabcolsep}{3pt}
\begin{minipage}[t]{0.485\linewidth}
\vspace{0pt}
\begin{tabular*}{\linewidth}{@{\extracolsep{\fill}}lrrr@{}}
\toprule
Width & Scratch & Generated & $\Delta$ \\
\midrule
\multicolumn{4}{@{}l}{\textbf{kr-vs-kp}\quad $73\to32\to32\to2$}\\
$0.5\times$ & $\mstd{97.38}{0.72}$ & $\mstd{96.02}{1.62}$ & $-1.36$ \\
$1\times$ & $\mstd{98.61}{0.51}$ & $\mstd{97.62}{0.60}$ & $-0.99$ \\
$1.5\times$ & $\mstd{99.13}{0.39}$ & $\mstd{97.13}{0.94}$ & $-2.00$ \\
\addlinespace[3pt]
\multicolumn{4}{@{}l}{\textbf{letter}\quad $16\to32\to32\to26$}\\
$0.5\times$ & $\mstd{79.54}{0.80}$ & $\mstd{54.38}{3.51}$ & $-25.16$ \\
$1\times$ & $\mstd{87.07}{0.50}$ & $\mstd{73.26}{1.82}$ & $-13.81$ \\
$1.5\times$ & $\mstd{90.45}{0.35}$ & $\mstd{73.96}{1.66}$ & $-16.49$ \\
\addlinespace[3pt]
\multicolumn{4}{@{}l}{\textbf{balance-scale}\quad $4\to16\to3$}\\
$0.5\times$ & $\mstd{95.84}{1.82}$ & $\mstd{96.79}{1.32}$ & $+0.95$ \\
$1\times$ & $\mstd{96.96}{1.10}$ & $\mstd{96.83}{1.15}$ & $-0.13$ \\
$1.5\times$ & $\mstd{97.48}{0.91}$ & $\mstd{96.95}{1.18}$ & $-0.53$ \\
\addlinespace[3pt]
\multicolumn{4}{@{}l}{\textbf{mfeat-factors}\quad $216\to32\to10$}\\
$0.5\times$ & $\mstd{96.17}{0.45}$ & $\mstd{94.78}{0.88}$ & $-1.39$ \\
$1\times$ & $\mstd{96.65}{0.37}$ & $\mstd{96.08}{0.48}$ & $-0.57$ \\
$1.5\times$ & $\mstd{96.80}{0.31}$ & $\mstd{96.39}{0.46}$ & $-0.41$ \\
\addlinespace[3pt]
\multicolumn{4}{@{}l}{\textbf{mfeat-fourier}\quad $76\to64\to10$}\\
$0.5\times$ & $\mstd{83.22}{1.00}$ & $\mstd{83.15}{1.37}$ & $-0.07$ \\
$1\times$ & $\mstd{85.14}{0.87}$ & $\mstd{84.29}{1.03}$ & $-0.85$ \\
$1.5\times$ & $\mstd{85.86}{0.78}$ & $\mstd{84.29}{0.84}$ & $-1.57$ \\
\addlinespace[3pt]
\multicolumn{4}{@{}l}{\textbf{breast-w}\quad $9\to64\to32\to2$}\\
$0.5\times$ & $\mstd{96.31}{0.28}$ & $\mstd{95.48}{0.81}$ & $-0.83$ \\
$1\times$ & $\mstd{96.28}{0.31}$ & $\mstd{96.21}{0.36}$ & $-0.07$ \\
$1.5\times$ & $\mstd{95.86}{0.48}$ & $\mstd{96.41}{0.12}$ & $+0.55$ \\
\addlinespace[3pt]
\multicolumn{4}{@{}l}{\textbf{mfeat-karhunen}\quad $64\to16\to8\to10$}\\
$0.5\times$ & $\mstd{59.51}{7.43}$ & $\mstd{64.63}{6.25}$ & $+5.12$ \\
$1\times$ & $\mstd{89.04}{2.57}$ & $\mstd{87.02}{3.00}$ & $-2.02$ \\
$1.5\times$ & $\mstd{93.09}{0.95}$ & $\mstd{90.54}{1.39}$ & $-2.55$ \\
\addlinespace[3pt]
\multicolumn{4}{@{}l}{\textbf{mfeat-morphological}\quad $6\to64\to32\to10$}\\
$0.5\times$ & $\mstd{71.55}{0.99}$ & $\mstd{72.01}{1.78}$ & $+0.46$ \\
$1\times$ & $\mstd{72.62}{0.84}$ & $\mstd{72.45}{1.25}$ & $-0.17$ \\
$1.5\times$ & $\mstd{73.87}{0.97}$ & $\mstd{71.69}{0.92}$ & $-2.18$ \\
\addlinespace[3pt]
\multicolumn{4}{@{}l}{\textbf{mfeat-zernike}\quad $47\to24\to10$}\\
$0.5\times$ & $\mstd{73.41}{1.65}$ & $\mstd{75.97}{1.81}$ & $+2.56$ \\
$1\times$ & $\mstd{77.50}{1.26}$ & $\mstd{77.28}{1.18}$ & $-0.22$ \\
$1.5\times$ & $\mstd{79.15}{1.12}$ & $\mstd{77.67}{1.13}$ & $-1.48$ \\
\addlinespace[3pt]
\multicolumn{4}{@{}l}{\textbf{cmc}\quad $24\to40\to20\to3$}\\
$0.5\times$ & $\mstd{54.94}{1.64}$ & $\mstd{55.71}{1.83}$ & $+0.77$ \\
$1\times$ & $\mstd{58.15}{1.28}$ & $\mstd{57.87}{1.37}$ & $-0.28$ \\
$1.5\times$ & $\mstd{58.82}{1.22}$ & $\mstd{58.12}{1.22}$ & $-0.70$ \\
\bottomrule
\end{tabular*}
\end{minipage}
\hfill
\begin{minipage}[t]{0.485\linewidth}
\vspace{0pt}
\begin{tabular*}{\linewidth}{@{\extracolsep{\fill}}lrrr@{}}
\toprule
Width & Scratch & Generated & $\Delta$ \\
\midrule
\multicolumn{4}{@{}l}{\textbf{blood-transfusion}\quad $4\to32\to2$}\\
$0.5\times$ & $\mstd{76.55}{1.24}$ & $\mstd{77.33}{1.06}$ & $+0.78$ \\
$1\times$ & $\mstd{77.49}{0.93}$ & $\mstd{77.46}{0.92}$ & $-0.03$ \\
$1.5\times$ & $\mstd{78.17}{0.74}$ & $\mstd{77.64}{0.89}$ & $-0.53$ \\
\addlinespace[3pt]
\multicolumn{4}{@{}l}{\textbf{banknote-auth}\quad $4\to32\to2$}\\
$0.5\times$ & $\mstd{94.27}{2.13}$ & $\mstd{95.67}{3.27}$ & $+1.40$ \\
$1\times$ & $\mstd{96.38}{0.89}$ & $\mstd{96.24}{0.85}$ & $-0.14$ \\
$1.5\times$ & $\mstd{96.88}{0.41}$ & $\mstd{96.37}{0.71}$ & $-0.51$ \\
\addlinespace[3pt]
\multicolumn{4}{@{}l}{\textbf{hayes-roth}\quad $4\to16\to3$}\\
$0.5\times$ & $\mstd{65.38}{6.46}$ & $\mstd{79.41}{6.93}$ & $+14.03$ \\
$1\times$ & $\mstd{81.52}{4.21}$ & $\mstd{82.22}{4.02}$ & $+0.70$ \\
$1.5\times$ & $\mstd{81.78}{3.78}$ & $\mstd{81.37}{3.48}$ & $-0.41$ \\
\addlinespace[3pt]
\multicolumn{4}{@{}l}{\textbf{BNG(cmc)}\quad $24\to40\to20\to3$}\\
$0.5\times$ & $\mstd{57.57}{0.28}$ & $\mstd{54.52}{1.44}$ & $-3.05$ \\
$1\times$ & $\mstd{57.88}{0.29}$ & $\mstd{57.10}{0.48}$ & $-0.78$ \\
$1.5\times$ & $\mstd{57.83}{0.30}$ & $\mstd{57.18}{0.49}$ & $-0.65$ \\
\addlinespace[3pt]
\multicolumn{4}{@{}l}{\textbf{heart-c}\quad $25\to32\to2$}\\
$0.5\times$ & $\mstd{78.93}{3.46}$ & $\mstd{76.97}{3.76}$ & $-1.96$ \\
$1\times$ & $\mstd{80.80}{2.53}$ & $\mstd{81.39}{2.12}$ & $+0.59$ \\
$1.5\times$ & $\mstd{81.34}{2.11}$ & $\mstd{80.56}{2.60}$ & $-0.78$ \\
\addlinespace[3pt]
\multicolumn{4}{@{}l}{\textbf{heart-h}\quad $25\to32\to2$}\\
$0.5\times$ & $\mstd{75.61}{4.03}$ & $\mstd{79.15}{3.83}$ & $+3.54$ \\
$1\times$ & $\mstd{77.99}{3.28}$ & $\mstd{78.34}{3.69}$ & $+0.35$ \\
$1.5\times$ & $\mstd{80.42}{3.62}$ & $\mstd{77.85}{3.27}$ & $-2.57$ \\
\addlinespace[3pt]
\multicolumn{4}{@{}l}{\textbf{ozone-level-8hr}\quad $72\to32\to2$}\\
$0.5\times$ & $\mstd{95.15}{0.31}$ & $\mstd{95.27}{0.39}$ & $+0.12$ \\
$1\times$ & $\mstd{95.46}{0.27}$ & $\mstd{95.40}{0.32}$ & $-0.06$ \\
$1.5\times$ & $\mstd{95.56}{0.26}$ & $\mstd{95.42}{0.28}$ & $-0.14$ \\
\addlinespace[3pt]
\multicolumn{4}{@{}l}{\textbf{phoneme}\quad $5\to32\to2$}\\
$0.5\times$ & $\mstd{78.15}{0.68}$ & $\mstd{78.62}{0.88}$ & $+0.47$ \\
$1\times$ & $\mstd{78.88}{0.67}$ & $\mstd{78.78}{0.72}$ & $-0.10$ \\
$1.5\times$ & $\mstd{79.30}{0.71}$ & $\mstd{78.82}{0.58}$ & $-0.48$ \\
\addlinespace[3pt]
\multicolumn{4}{@{}l}{\textbf{solar-flare (5 classes)}\quad $34\to24\to12\to5$}\\
$0.5\times$ & $\mstd{78.25}{2.59}$ & $\mstd{70.40}{5.83}$ & $-7.85$ \\
$1\times$ & $\mstd{75.44}{2.73}$ & $\mstd{74.37}{3.51}$ & $-1.07$ \\
$1.5\times$ & $\mstd{75.92}{2.18}$ & $\mstd{75.52}{2.93}$ & $-0.40$ \\
\addlinespace[3pt]
\multicolumn{4}{@{}l}{\textbf{solar-flare (2 classes)}\quad $31\to32\to2$}\\
$0.5\times$ & $\mstd{83.01}{0.57}$ & $\mstd{82.97}{0.58}$ & $-0.04$ \\
$1\times$ & $\mstd{83.09}{0.59}$ & $\mstd{83.23}{0.60}$ & $+0.14$ \\
$1.5\times$ & $\mstd{83.00}{0.55}$ & $\mstd{83.05}{0.59}$ & $+0.05$ \\
\midrule
\multicolumn{4}{@{}l}{\textbf{Mean across all datasets}}\\
$0.5\times$ & $79.54$ & $78.96$ & $-0.58$ \\
$1\times$ & $83.15$ & $82.17$ & $-0.98$ \\
$1.5\times$ & $84.04$ & $82.35$ & $-1.69$ \\
\bottomrule
\end{tabular*}
\end{minipage}
\end{table}

\end{document}